\documentclass{article}
\usepackage{ijcai23}

\usepackage{times}
\usepackage{soul}
\usepackage{url}
\usepackage[hidelinks]{hyperref}
\usepackage[utf8]{inputenc}
\usepackage[small, justification=centering]{caption}
\usepackage{graphicx}
\usepackage{amsmath}
\usepackage{amsthm}
\usepackage{booktabs}
\usepackage[export]{adjustbox}
\usepackage{algorithm}
\usepackage{algorithmic}
\usepackage[switch]{lineno}
\usepackage{amsfonts}       
\usepackage{nicefrac}       
\usepackage{microtype}      
\usepackage{acronym}
\usepackage{bmpsize}
\usepackage{paralist}
\usepackage{todonotes}
\usepackage{mathtools}
\usepackage{scalerel}
\usepackage{xcolor}
\usepackage{amssymb}
\usepackage{multirow}
\usepackage{makecell}
\usepackage{caption}
\usepackage{subcaption}
\usepackage{varwidth}
\usepackage{float}
\usepackage{hyperref}
\usepackage{listings,lstautogobble}
\usepackage{calc}
\newlength\myheight
\newlength\mydepth
\settototalheight\myheight{Xygp}
\newcommand*\inlinegraphics[1]{%
  \settototalheight\myheight{Xygp}%
  \settodepth\mydepth{Xygp}%
  \raisebox{-\mydepth}{\includegraphics[height=\myheight]{#1}}%
}

\newtheorem{example}{Example}

\DeclareMathOperator*{\argmax}{\arg\max}
\DeclareMathOperator*{\argmin}{\arg\min}
\newcommand{\asp}[1]{\mbox{$\mathtt{#1}$}}

\acrodef{nsl}[NSIL]{Neuro-Symbolic Inductive Learner}
\acrodef{asp}[ASP]{Answer Set Programming}
\acrodef{las}[LAS]{Learning from Answer Sets}
\acrodef{ilp}[ILP]{Inductive Logic Programming}
\acrodef{ai}[AI]{Artificial Intelligence}
\acrodef{cnn}[CNN]{Convolutional Neural Network}
\acrodef{wcdpi}[WCDPI]{Weighted Context-Dependant Partial Interpretation}
\acrodef{fnr}[FNR]{False Negative Rate}
\acrodef{bnn}[BNN]{Binary Neural Network}
\acrodef{cbm}[CBM]{Concept Bottleneck Model}

\title{Neuro-Symbolic Learning of Answer Set Programs from Raw Data}

\author{
Daniel Cunnington$^{1,2}$\and
Mark Law$^3$\and
Jorge Lobo$^4$\And
Alessandra Russo$^2$\\
\affiliations
$^1$IBM Research Europe\\
$^2$Imperial College London\\
$^3$ILASP Limited\\
$^4$ICREA-Universitat Pompeu Fabra\\
\emails
dancunnington@uk.ibm.com, mark@ilasp.com, jorge.lobo@upf.edu, a.russo@imperial.ac.uk
}

\begin{document}

\maketitle

\begin{abstract}
One of the ultimate goals of Artificial Intelligence is to assist humans in complex decision making. A promising direction for achieving this goal is Neuro-Symbolic AI, which aims to combine the interpretability of symbolic techniques with the ability of deep learning to learn from raw data. However, most current approaches require manually engineered symbolic knowledge, and where end-to-end training is considered, such approaches are either restricted to learning definite programs, or are restricted to training binary neural networks. In this paper, we introduce Neuro-Symbolic Inductive Learner (NSIL), an approach that trains a general neural network to extract latent concepts from raw data, whilst learning symbolic knowledge that maps latent concepts to target labels. The novelty of our approach is a method for biasing the learning of symbolic knowledge, based on the in-training performance of both neural and symbolic components. We evaluate NSIL on three problem domains of different complexity, including an NP-complete problem. Our results demonstrate that NSIL learns expressive knowledge, solves computationally complex problems, and achieves state-of-the-art performance in terms of accuracy and data efficiency. Code and technical appendix: \url{https://github.com/DanCunnington/NSIL}
\end{abstract}

\section{Introduction}
Within \ac{ai}, one of the ultimate goals is to assist humans in complex decision making, a key challenge in multiple industries such as healthcare, automated maintenance, and security \cite{lyn2019opportunities,sitton2019neuro,han2021unifying}. 
Neuro-Symbolic \ac{ai} aims to address this challenge by combining the best features of both deep learning and symbolic reasoning techniques \cite{GarcezGLSST19,ijcai2020-0688}. 
For example, many existing approaches improve the training of a neural network using a given symbolic knowledge base \cite{deepproblog,dai2019bridging,riegel2020logical,neurasp,Badreddine2022}, and systems that learn symbolic knowledge can utilise pre-trained neural networks to handle raw input data \cite{evans2018learning,cunnington2023ffnsl}. However, the assumption that either the neural or symbolic component is given is not practical in many real-world situations, as there may be no pre-trained neural network(s) available (e.g., domain experts are time constrained and can't label vast quantities of raw data), or the symbolic knowledge may be unknown, and therefore needs to be learned. Recent work aims to lift this assumption by training both neural and symbolic components in an end-to-end fashion \cite{ijcai2021-254,EVANS2021103521}. However, \cite{ijcai2021-254} lacks the expressivity required to efficiently represent solutions to many common-sense learning and reasoning tasks \cite{dantsin2001complexity}, and \cite{EVANS2021103521} is restricted to training \acp{bnn} only.

In this paper, we introduce \textit{\ac{nsl}}, which trains a \textit{general} neural network to classify latent concepts from raw data, whilst learning an expressive and interpretable knowledge base that solves computationally \textit{complex} problems. \ac{nsl} only requires a target label for each training data point, and no latent concept labels are given. \ac{nsl} uses a state-of-the-art symbolic learner to learn a logic program in the language of \ac{asp} \cite{gelfond2014knowledge}, a highly expressive form of knowledge representation which can efficiently represent solutions to problems of computational complexity greater than P \cite{dantsin2001complexity}. A general neural network architecture is trained by \textit{reasoning} over the learned knowledge, where the latent concept values that the network should output in order to obtain each target label are identified. We use NeurASP to perform such computation \cite{neurasp}, and seamlessly integrate symbolic learning by creating \textit{corrective examples} for the symbolic learner. A specific mechanism for weighting these examples is proposed, which biases the symbolic learner to either refine or retain the learned knowledge, balancing \textit{exploration} with \textit{exploitation} of the symbolic search space. 

We evaluate \ac{nsl} on three problem domains: (1) \textit{Cumulative Arithmetic}, where the learned programs perform the cumulative addition or product over sequences of MNIST digit images. (2) \textit{Two-Digit Arithmetic}, where the learned programs include arithmetic and logical operations over two MNIST digit images. (3) the \textit{Hitting Set} problem, which includes the standard decision problem and a variant that remains NP-complete \cite{karp1972reducibility}. In this domain, the learned programs not only solve the decision problem, but can also generate \textit{all} the hitting sets of a given collection. In all domains, the input is a set of training data points, each containing a sequence of raw images together with a target label. Also, the full space of possible latent concepts is assumed to be known, together with some structural properties of the search space for the symbolic learner. The output is a trained neural network capable of classifying raw images into latent concepts, and an \ac{asp} program that maps latent concepts into target labels. 

Finally, we compare \acp{nsl} performance with purely neural baselines, and where appropriate, $Meta_{Abd}$ \cite{ijcai2021-254}, the only existing system that trains a general neural network and a symbolic learner. The results show that \ac{nsl}: (1) Outperforms the baselines in terms of overall accuracy and data efficiency. (2) Learns expressive knowledge to represent solutions to an NP-complete problem, which would be difficult to express with any of the other systems. (3) Requires significantly fewer data points than the neural baselines. (4) Trains the neural network to predict latent concepts with an accuracy comparable to fully supervised training, and (5) achieves comparable performance to approaches that either assume the latent concept labels are given, or the symbolic component is given and only the neural network needs to be trained.


\section{Related Work}
Many Neuro-Symbolic integrations have been proposed in the literature to enhance data efficiency, transferability and interpretability of neural networks using symbolic techniques \cite{Besold2017}. Some approaches inject symbolic knowledge directly into neural architectures \cite{riegel2020logical,Badreddine2022}, whereas others preserve a clear distinction between neural and symbolic components \cite{deepproblog,neurasp,aspis2022embed2sym,Deeppsl}. The main drawback of these approaches is that they require a complete and manually engineered symbolic knowledge base. On the other hand, symbolic learning systems \cite{muggleton1994inductive,Corapi10,Muggleton13,Law2018thesis,law2020fastlas} are capable of learning interpretable knowledge in a data efficient manner, but are restricted to learning from structured data, even when differentiable training methods are used \cite{payani2019inductive,Shindo_Nishino_Yamamoto_2021,sen2021neurosymbolic}. Therefore, pre-trained neural networks are often required when learning from raw data \cite{evans2018learning,ferreira2022looking,cunnington2023ffnsl}. In contrast to all of the aforementioned approaches, our method performs joint Neuro-Symbolic learning, by training a neural network and learning symbolic knowledge simultaneously, which significantly reduces the amount of engineering and labelling required. 

\ac{nsl} is therefore closely related to the existing approaches that also train both neural and symbolic components \cite{ijcai2021-254,EVANS2021103521,dsl}. $Meta_{Abd}$ \cite{ijcai2021-254} extends \cite{dai2019bridging} by using abduction and induction to jointly train a neural network and learn symbolic knowledge from raw data. Specifically, the symbolic knowledge learned by $Meta_{Abd}$ is used to abduce possible latent concept values for training the neural network. These values are pruned further and a unique value is selected for each image within an input sequence, using the neural network confidence. As the authors note, the network is therefore vulnerable to becoming stuck in a local optima if the abductive space is very dense. 
During the early stages of training, the network predicts with a more equal distribution, which reduces the likelihood of selecting the correct latent concept values. In contrast, \ac{nsl} relies on a semantic loss function \cite{neurasp}, and instead trains the network with a set of possible values. This avoids incorrectly pruning the abductive space and the network is able to escape local optima.

Also, the symbolic learner used by $Meta_{Abd}$ is restricted in the expressivity of knowledge that can be learned \cite{Law2018thesis}. As \ac{nsl} learns \ac{asp} programs, we can represent common-sense knowledge involving defaults, exceptions, constraints and choice \cite{gelfond2014knowledge}, all of which are outside the scope of $Meta_{Abd}$. Using \ac{asp} also enables our learned programs to have multiple answer sets (multiple models), whereas $Meta_{Abd}$ is restricted to learning definite programs that accept only one model. These features enable \ac{nsl} to learn knowledge that efficiently represents solutions to complex, NP-complete problems, which would be difficult to represent using $Meta_{Abd}$.

The approach in \cite{EVANS2021103521} encodes the neural network into an \ac{asp} program, which enables joint neural network training and symbolic learning within \ac{asp}. However, \ac{asp} does not support continuous arithmetic, which restricts \cite{EVANS2021103521} to using \acp{bnn}. \acp{bnn} are an active area of research, but are currently limited to relatively small network architectures. Due to our modular approach, \ac{nsl} does not place such restrictions on the neural network architecture, and can take advantage of GPU-accelerated hardware and software. Finally, \cite{dsl} learns a perception and reasoning function, but the reasoning function is a look-up table that simply maps symbolic inputs to outputs. In contrast, \ac{nsl} is able to learn complex reasoning functions in the form of \ac{asp} programs, which are more expressive than a lookup-table, and can generalise beyond the specific task used during training.
\section{Neuro-Symbolic Inductive Learner}

\begin{figure*}[t]
    \centering
    \begin{tabular}[t]{cc}
    \begin{subfigure}[c]{0.5\textwidth}
        \centering  
        \includegraphics[keepaspectratio, width=\textwidth]{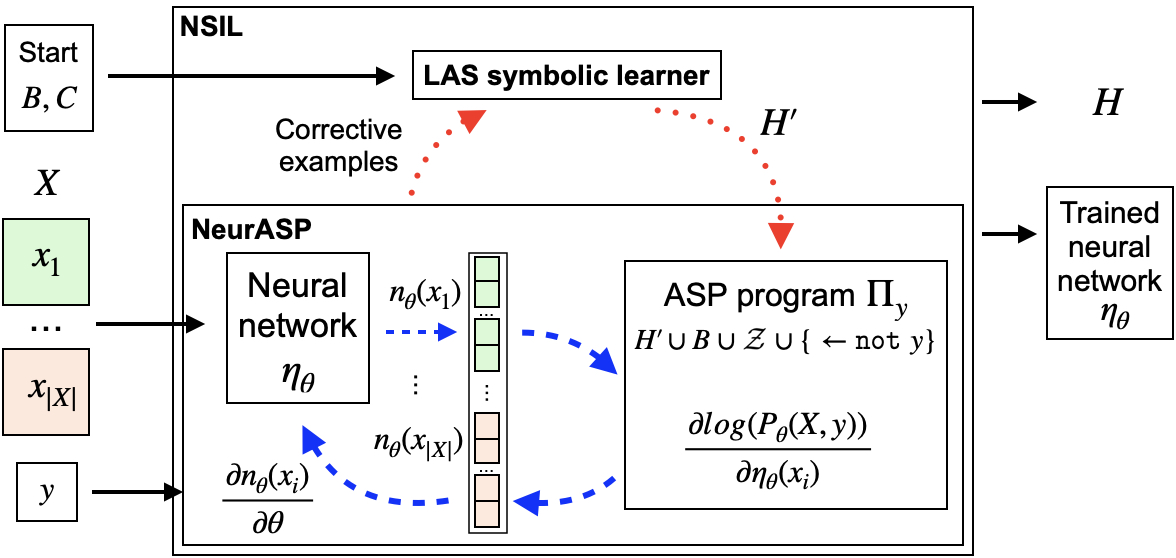}
        \caption{Learning}
    \end{subfigure}
    \hfill
    \hspace{2em}
    \hfill
    \begin{tabular}{c}
        \begin{subfigure}[t]{0.3\textwidth}
            \centering
            \includegraphics[keepaspectratio, width=\textwidth]{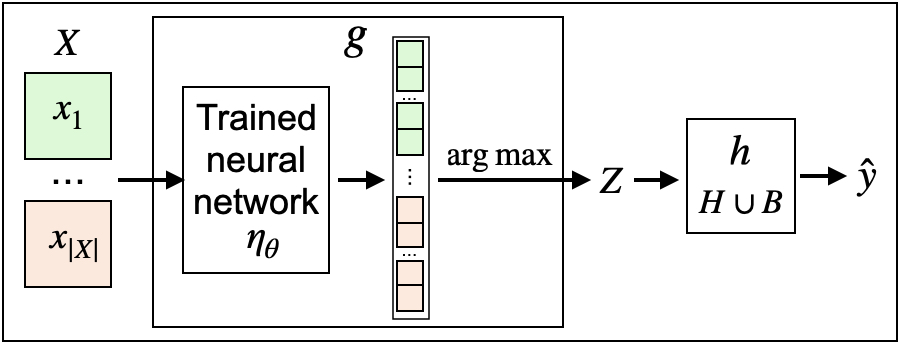}
            \caption{Inference}\label{fig:inference}
        \bigskip
        \end{subfigure}\\
        \begin{subfigure}[t]{0.3\textwidth}
            \centering
            \includegraphics[keepaspectratio, width=\textwidth]{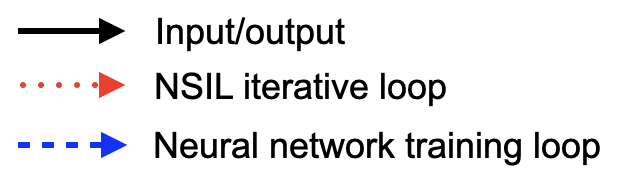}
        \end{subfigure}
    \end{tabular}\\
    \end{tabular}
    \caption{(a) \ac{nsl} learning with a single data point $\langle X,y \rangle \in D$. (b) \ac{nsl} inference over a single input $X$.}
    \label{fig:arch}
\end{figure*}

\subsection{Problem Setting}
We decompose the task of learning complex knowledge from raw data into a perception function $g:\mathcal{X}\rightarrow\mathcal{Z}$ and a reasoning function $h:\mathcal{Z}^{a,b}\rightarrow \mathcal{Y}$. $\mathcal{X}$ is the space of raw inputs (e.g., images), $\mathcal{Z}$ is the space of latent concept values, and $\mathcal{Y}$ is the space of target labels. $h$ is defined over sequences of latent concept values, with (possibly) varying length. Therefore, $\mathcal{Z}^{a,b}$ is the set of all possible sequences of $\mathcal{Z}$ with length $\geq a$ and $ \leq b$. A latent concept is a tuple $C=\langle n, \mathcal{Z}\rangle$, where $n$ is the name for the latent concept. In this paper, we assume w.l.o.g. there is a single latent concept for a given task. Supporting multiple latent concepts could be achieved by partitioning $\mathcal{Z}$ into different classes, and assigning each class a different name with a different perception function. During training, we observe a dataset of samples $D=\{\langle X, y\rangle, \ldots\}$, where $X$ is a sequence of raw inputs $x\in\mathcal{X}$, and $y\in\mathcal{Y}$ is a target label. We assume the latent concept $C$ is given as background knowledge, but crucially, the samples in $D$ are not annotated with latent concept labels.

\begin{example}\label{ex:mnist_addition}
Consider the MNIST Addition task from \cite{deepproblog}. The goal is to learn a perception function $g$ that classifies raw MNIST images into a latent concept $C= \langle \asp{digit}, \{0\ .. \ 9\}\rangle$, and a reasoning function $h$ that returns the sum of two latent concept values, where $\mathcal{Y} = \{0\ ..\ 18\}$. $D$ contains training samples of two MNIST images, and the label indicates the sum, e.g., $X=[\scalerel*{\includegraphics{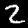}}{)},\text{\scalerel*{\includegraphics{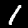}}{)}}]$ and $y=3$.
\end{example}

To combine the benefits of neural perception with symbolic reasoning, we implement $g$ using a neural network, and $h$ using a first-order logic program in the language of \ac{asp}. Jointly learning both $g$ and $h$ is challenging, as $h$ is not differentiable and cannot be learned using standard gradient-based optimisation. To learn $h$, we use a symbolic learner that explores a search space constructed from symbolic relations given in a domain knowledge $B$. Therefore, assuming $B$ and $C$ are given as input, the objective of \ac{nsl} is to learn $g$ and $h$, s.t. $\forall \langle X,y \rangle \in D$, $h([g(x_{i}) :x_{i}\in X]) = y$. Let us now describe how $g$ and $h$ are learned within the \ac{nsl} architecture.

\subsection{Neural and Symbolic Components}\label{sec:method:components}
The \ac{nsl} architecture is presented in Figure \ref{fig:arch}. During learning, \ac{nsl} trains a neural network using knowledge learned by a symbolic learner. Corrective examples then bias the symbolic learner to refine or retain the learned knowledge based on the current system performance. The neural network is then updated again, as part of an iterative training process. During inference, the trained neural network forms the perception function $g$, and the learned knowledge forms the reasoning function $h$, to enable target predictions $\hat{y}\in \mathcal{Y}$ from unseen raw data. Let us now describe the optimisation of the neural and symbolic components.

\textbf{Symbolic learner}. The symbolic component is a \ac{las} symbolic learner \cite{LawRB19}. The input is a domain knowledge $B$ that contains any (optional) background knowledge, and relations used to construct a search space $\mathcal{H}$. The symbolic learner also accepts a set of examples $E$ as input, where in \ac{nsl}, each example associates a sequence of latent concept values $Z \in \mathcal{Z}^{a,b}$ with a target label $y\in\mathcal{Y}$. The output is a learned knowledge base $H\in \mathcal{H}$ which is a first-order logic program in the language of \ac{asp}. $H$, together with $B$, forms the reasoning function $h=H\cup B$ that maps sequences of latent concept values to target labels. We configure each example $e\in E$ to represent the label either positively (i.e., $y$ should be output for $Z$), or negatively ($y$ should \textit{not} be output for $Z$). $H$ is said to \textit{cover} an example, if $H \cup B$ outputs a desired label for the sequence of latent concept values (where a desired label depends on the positive or negative example configuration). Also, $e$ is weighted with an integer penalty $W(e)$, paid by $H$ for leaving the example uncovered. Details of how the examples are constructed are presented in Section \ref{sec:method:corrective_ex}. During learning, we add constraints to $B$ to ensure $H$ is guaranteed to output each target label we observe in $D$, i.e., $\forall \langle X,y\rangle \in D$, there exists a sequence of latent concept values $Z \in \mathcal{Z}^{a,b}$ s.t. $h(Z)=y$, where $h=H\cup B$. Also, $H$ minimises a scoring function, based on the length of $H$ (i.e., the number of literals), and the coverage over the examples $E$. Let us assume the set of examples uncovered by $H$ is denoted by $UNCOV(H,(B,E))$. The score of $H$ is given by $score(H,(B,E)) = \vert H \vert +  \sum_{e\in UNCOV(H,(B,E))}{W(e)}$. A \ac{las} symbolic learner solves the following optimisation:

\begin{equation}
\label{lasequation}
H^{*} = \argmin_{H\in \mathcal{H}}\; [ score(H, (B,E)) ]
\end{equation}
This can be equivalently interpreted as jointly maximising the generality of $H$ (i.e., the most compressed \ac{asp} program), and the coverage of the examples in $E$.

\textbf{Neural network}. For the neural component, we define a neural network $\eta_{\theta}:\mathcal{X}\rightarrow [0,1]^{\vert \mathcal{Z}\vert}$ that maps a raw input $x\in\mathcal{X}$ to a probability distribution over the possible latent concept values $\mathcal{Z}$. $\theta$ denotes the neural network parameters which need to be learned. During inference, the perception function $g$ returns the latent concept value with the maximum probability using the standard $\argmax$ function, i.e., $g(x) = \argmax_{z\in\mathcal{Z}}(\eta_{\theta}(x)[z])$. As there are no latent concept labels available, the neural network is trained using a symbolic knowledge $H$, alongside $B$ and the target label $y$. Recall that the \textit{space} of possible latent concept values $\mathcal{Z}$ is given as background knowledge. As $H$ is in the language of \ac{asp}, we use NeurASP \cite{neurasp} to optimise a semantic loss function \cite{Xu18} for neural network training. Informally, for each training data point, the neural network is trained to predict the latent concept values that result in the target label, given $H$ and $B$. 

\begin{example}\label{ex:mnist_neurasp}
   Following Example \ref{ex:mnist_addition}, let us assume $H$ is the correct addition rule. For $X=[\scalerel*{\includegraphics{figures/digits/2.png}}{)},\text{\scalerel*{\includegraphics{figures/digits/1.png}}{)}}], \ y=3$, the neural network would be trained to output either $[0,3]$, $[1,2]$, $[2,1]$, or $[3,0]$ for these images. Note that each of these latent concept values receives a gradient update, and the network learns to distinguish individual digits by observing multiple data points.
\end{example}

Formally, consider a training sample $\langle X,y\rangle \in D$. We define a matrix $M_{\theta}(X)$ which contains all neural network outputs for each image within $X$. $M_{\theta}(X)[i,z]$ denotes the neural network output for image $x_{i}\in X$ and latent concept value $z$. E.g., following Example \ref{ex:mnist_neurasp}, $M_{\theta}(X)[1,2]$ corresponds to the probability of the first image $\scalerel*{\includegraphics{figures/digits/2.png}}{)}$ having latent concept value 2. We also define an \ac{asp} program $\Pi_y = H \cup B  \cup \mathcal{Z} \cup \{\leftarrow \asp{not} \; y\}$, with answer sets\footnote{The answer sets correspond to the solutions of the program. See \cite{gelfond2014knowledge} for an overview of \ac{asp}.} denoted $AS(\Pi_{y})$. Each answer set contains a sequence of latent concept values $Z$ that output label $y$, given $H$ and $B$. We define the probability of each $Z\in AS(\Pi_y)$ in terms of the neural network output: $P_{\theta,X}(Z)= \prod_{z_{i} \in Z}{M_{\theta}(X)[i,z_{i}]}$. The probability of an $\langle X,y\rangle$ training sample is then defined as $P_{\theta}(X,y) = \sum_{Z \in AS(\Pi_{y})}{P_{\theta,X}(Z)}$, and the neural network is trained to optimise:

\begin{equation}
\label{neurasp_eqn}
\theta^{*} = \argmax_{\theta}\; [ \sum_{\langle X,y\rangle \in D}{log(P_{\theta}(X,y))} ]
\end{equation}

Intuitively, this can be interpreted as training the neural network to maximise the probability of the latent concept values that output each downstream label, given $H$ and $B$. For further details, we refer the reader to \cite{neurasp}.\footnote{In this paper, we assume each data point has a single $y$ label. To simplify notation, we have presented a modified version of NeurASP that also assumes a single label. Generalising \ac{nsl} to multi-label tasks is left as future work.} Now we have outlined the neural and symbolic components, we can introduce our novel method of integration. Recall that the symbolic learner learns a knowledge base which is used to train the neural network. An immediate question is; what happens if the learned knowledge is incorrect, as that may lead to a sub-optimal neural network? To tackle this problem, \ac{nsl} adopts an iterative training procedure where both the symbolic learner and the neural network are updated on each iteration. This integration relies upon a set of weighted \textit{corrective examples} for the symbolic learner, that encourages the current knowledge to be refined or retained on the next iteration. Let us now outline how the corrective examples are structured and how their weights are updated on each iteration.

\subsection{Corrective Examples}\label{sec:method:corrective_ex}
A corrective example $e_{Z,y}$ associates a sequence of latent concept values $Z \in \mathcal{Z}^{a,b}$ with a target label $y\in\mathcal{Y}$. Each example is configured to represent the label either positively (i.e., $y$ should be output for $Z$, denoted $e_{Z,y}^{\text{pos}}$), or negatively ($y$ should \textit{not} be output for $Z$, denoted $e_{Z,y}^{\text{neg}}$). Also, recall that each example is weighted with a penalty, paid if the example is left \textit{uncovered}. A pair of corrective examples $\langle e_{Z,y}^{\text{pos}},e_{Z,y}^{\text{neg}} \rangle$ represents both positive and negative cases for a given $Z,y$ combination. In \ac{nsl}, the set of symbolic examples $E$ contains a pair of corrective examples for all possible $Z,y$ combinations, and initially, all of the examples have equal weight set to 0. This means there is no bias given to the symbolic learner as to which examples should be covered. We restrict the weights to the interval $[0,100]$, and the goal is to maximise the weights of the positive examples that contain correct combinations of latent concept values with target labels, whilst minimising the weights of the corresponding negative examples.

On each iteration, the neural network is trained with a candidate knowledge base, denoted $H^{\prime}$. The corrective example weights are then updated, using two sources of information: (1) The overall performance when predicting training set labels $y$. (2) The neural network confidence when predicting $Z$. In each case, we calculate a weight value, and associate this with a specific $Z,y$ combination. To capture the overall performance, we obtain $\hat{y}$ predictions with a forward pass over the training set using both neural and symbolic components. We then calculate the \ac{fnr} of each label $y\in\mathcal{Y}$, which depends on the current $H^{\prime}$ and current neural network parameters, denoted $\theta^{\prime}$:

\begin{equation}\label{eq:fnr}
FNR_{H^{\prime},\theta^{\prime}}(y) = 100\times \left (1 -
\frac{TP_{H^{\prime},\theta^{\prime}}(y)}{TP_{H^{\prime},\theta^{\prime}}(y) + FN_{H^{\prime},\theta^{\prime}}(y)} \right )
\end{equation}

\noindent where $TP_{H^{\prime},\theta^{\prime}}(y)$ and $FN_{H^{\prime},\theta^{\prime}}(y)$ are the number of true positives and false negatives of label $y$ respectively. We multiply by 100 to map the \ac{fnr} into our $[0,100]$ weight range. Intuitively, if \ac{nsl} accurately predicts label $y$, $FNR_{H^{\prime},\theta^{\prime}}(y)$ will be close to 0, otherwise $FNR_{H^{\prime},\theta^{\prime}}(y)$ will be close to 100. To associate the \ac{fnr} with a $Z,y$ combination, we obtain all possible $Z$ that currently output label $y$ given the current $H^{\prime}$, by intercepting the answer sets of $\Pi_{y}$ within NeurASP (see Section \ref{sec:method:components}). We divide the \ac{fnr} by the number of answer sets of $\Pi_{y}$, i.e., $\frac{FNR_{H^{\prime},\theta^{\prime}}(y)}{\vert AS(\pi_{y})\vert}$, as there could be multiple $Z$ that lead to each $y$. E.g., in Example \ref{ex:mnist_neurasp} there are four sequences of latent concept values that lead to the label $y=3$, so the \ac{fnr} is shared between these possibilities.

To capture the neural network confidence, and associate it with a $Z,y$ combination, we compute an average confidence over all training data points with a label $y$, where the neural network predicts $Z$. Formally, let $D_{Z,y}$ be the set of $\langle X,y \rangle \in D$ with label $y$, where the neural network predicts the sequence of latent concept values $Z$ for the input sequence $X$, i.e., $[g(x_{i}) : x_{i} \in X]=Z$. Let us also define the probability of $X$ as the product of the maximal neural network confidence score for each input in the sequence: $P_{\theta^{\prime}}(X)=\prod_{\substack{x_{i} \in X}}{max(\eta_{\theta^{\prime}}(x_{i}))}$. The overall confidence for a $Z,y$ combination is defined as:

\begin{equation}\label{eq:network}
    CONF_{\theta^{\prime}}(Z,y) = \frac{100}{\vert D_{Z,y} \vert} \times \sum_{\langle X,y\rangle \in D_{Z,y}}{P_{\theta^{\prime}}(X)}
\end{equation}

Again, we multiply by 100 to map into our $[0,100]$ weight range. Given our two sources of information, we now outline how the corrective example weights are updated. On each iteration, we firstly set the weights of the $Z,y$ pairs that we have observed from the answer sets of $\Pi_y$ as follows:
\begin{align}
    W(e_{Z,y}^{\text{pos}}) &= 0\label{eq:pi_y_pos_weight}\\
    W(e_{Z,y}^{\text{neg}}) &= \frac{FNR_{H^{\prime},\theta^{\prime}}(y)}{\vert AS(\pi_{y})\vert}\label{eq:pi_y_neg_weight}
\end{align}

\noindent The negative example weight is set by the \ac{fnr}, as if the \ac{fnr} is large, this will encourage a different label for $Z$. The positive example weight is set to 0 because the \ac{fnr} may apply to multiple $Z,y$ pairs, and may not indicate any positive information for an individual pair. We then update the example weights using the neural network confidence to ensure positive information for an individual $Z,y$ pair is captured:

\begin{align}
    W(e_{Z,y}^{\text{pos}}) &= W(e_{Z,y}^{\text{pos}}) + \lambda CONF_{\theta^{\prime}}(Z,y)\label{eq:nn_pos_weights}\\
    W(e_{Z,y}^{\text{neg}}) &= W(e_{Z,y}^{\text{neg}}) - \lambda CONF_{\theta^{\prime}}(Z,y)\label{eq:nn_neg_weights}
\end{align}

\noindent where $\lambda \in [0,1]$ controls the effect of the $CONF_{\theta^{\prime}}$ update and is set to 1 in all of our experiments.\footnote{Please see Appendix A.1 for results with varying values of $\lambda$.} Example weights are allowed to persist across iterations, and in Equations \ref{eq:nn_pos_weights} and \ref{eq:nn_neg_weights}, the $W$ terms in the equation body could equal the weights set by Equations \ref{eq:pi_y_pos_weight} and \ref{eq:pi_y_neg_weight}, or the weights set on a previous iteration, if a $Z,y$ pair predicted by the neural network does not exist in the answer sets of $\Pi_{y}$. If there are no latent concept predictions of a certain $Z$, $CONF_{\theta^{\prime}}(Z,y)=0$. Finally, the weights are clipped to the range $[0,100]$.


\begin{example}\label{ex:fnr_correct}
Following Example \ref{ex:mnist_neurasp}, let $H^{\prime}$ be the correct addition rule, $y=3$ and $FNR_{H^{\prime},\theta^{\prime}}(y)=5$, indicating 95\% of training data points with label 3 have been predicted correctly. The answer sets of $\Pi_{y}$ are $[0,3]$, $[1,2]$, $[2,1]$, and $[3,0]$. Let us also assume that for training data points with label $3$, the correct $Z$ are predicted with 98\% confidence, and $\lambda=1$. Therefore, the positive corrective examples associated with these latent concept values all have weight 98, 
and the corresponding negative examples all have weight $(\frac{5}{4}) - 98 = -96.75$, which is clipped to 0. This encourages the symbolic learner to cover the positive examples on the next iteration, and leave the negative examples uncovered.
\end{example}

\begin{example}\label{ex:fnr_incorrect}
Conversely, let $H^{\prime}$ be an incorrect knowledge base, $y=3$, $\lambda=1$, and $FNR_{H^{\prime},\theta^{\prime}}(y)=95$ (indicating only 5\% of training data points with label 3 have been predicted correctly). Let us assume the answer sets of $\Pi_{y}$ are $[0,4]$, $[1,2]$, $[2,3]$, and, for all data points labelled 3, the neural network only predicts $[1,2]$ with 75\% confidence. Therefore:
$$W(e^{\text{pos}}_{[1,2],3})=75,\ \ \ \ \  W(e^{\text{pos}}_{[0,4],3}) = W(e^{\text{pos}}_{[2,3],3})=0,$$ 
$$W(e^{\text{neg}}_{[0,4],3})  =  W(e^{\text{neg}}_{[2,3],3}) = \frac{95}{3}=31.67,$$
$$W(^{\text{neg}}_{[1,2],3})= \left (\frac{95}{3}\right ) - 75 = -43.33 \ \text{ (which is clipped to 0).}$$

This encourages the symbolic learner to cover $e^{\text{pos}}_{[1,2],3}$, $e^{\text{neg}}_{[0,4],3}$ and $e^{\text{neg}}_{[2,3],3}$ on the next iteration.
\end{example}

For computational efficiency, example pairs with equal weights are not given to the symbolic learner. Also, weights are rounded to the nearest integer and shifted to the range $\{1\ ..\ 101 \}$ as the \ac{las} systems require integer weight penalties $> 0$. To summarise, \ac{nsl} solves Equations~\ref{lasequation} and \ref{neurasp_eqn} with the following steps:

\begin{enumerate}
    \item The symbolic learner learns an initial knowledge base $H^{\prime}$ that satisfies $B$ and covers each possible label in $\mathcal{Y}$. The neural network parameters $\theta^{\prime}$ are initialised randomly.
    \item The neural network is trained for 1 epoch on the training set $D$ to update $\theta^{\prime}$, using the \ac{asp} program $\Pi_{y}$ for each $\langle X,y\rangle \in D$. Note that $\Pi_{y}$ is constructed using $H^{\prime}$. 
    \item The weights of the corrective examples are updated (see Section~\ref{sec:method:corrective_ex}), and a new $H^{\prime}$ is learned.
    \item Steps 2-3 are repeated for a fixed number of iterations. \ac{nsl} outputs the learned neural network parameters $\theta$ and the learned knowledge base $H$.
\end{enumerate}

\section{Experiments}\label{sec:experiments}

Our experiments address the following questions: (1) Can \ac{nsl} train both neural and symbolic components, learning general, expressive and interpretable knowledge from raw data? (2) Is \ac{nsl} vulnerable to becoming stuck in local optima when training with limited data?
(3) Can \ac{nsl} learn the correct knowledge in large search spaces? (4) How does \ac{nsl} compare to the closest existing system $Meta_{Abd}$? (5) How does \ac{nsl} compare to end-to-end neural networks? Finally, (6) can \ac{nsl} learn expressive solutions to complex tasks that would be difficult to represent with other systems?

To answer these questions, we consider three problem domains; Cumulative Arithmetic, Two-Digit Arithmetic, and the Hitting Set problem. The Cumulative Arithmetic tasks follow \cite{ijcai2021-254} and enable a comparison to $Meta_{Abd}$, where \ac{nsl} achieves state-of-the-art performance, thus addressing question 4. To concretely demonstrate the contribution of \ac{nsl}, the other domains require learning more expressive knowledge that $Meta_{Abd}$ can't learn, including negation as failure, choice rules, and programs with multiple answer sets. We demonstrate \ac{nsl} trains both neural and symbolic components successfully, outperforms fully neural baselines, and learns the correct knowledge in large search spaces. This addresses questions 1, 3, and 5. Question 2 is also addressed here, where a reduced amount of training data is used in the Two-Digit Arithmetic domain, and \ac{nsl} is able to train successfully under these conditions. Question 6 is addressed in the Hitting Set domain, where \ac{nsl} is able to solve an NP-complete problem and genererate \textit{all} the hitting sets of a given collection, despite learning from binary labels. 

\textbf{Baselines.}
Aside from the comparison to $Meta_{Abd}$, we compare to \acp{cbm} \cite{koh2020concept}, which are end-to-end neural networks that model both perception and reasoning components. We ensure the perception network is the same as in \ac{nsl}, but the reasoning network is a multi-layer perceptron instead of a symbolic learner. We train two variants end-to-end (i.e., no concept labels are used), therefore using the same weak supervision as \ac{nsl}. One variant (denoted \textit{\ac{cbm}}) uses a linear layer as output from the perception network, and the second variant (denoted \textit{\ac{cbm}-S}) uses a softmax layer, replicating more closely the perception network used in \ac{nsl}. For the Two-Digit Arithmetic tasks, we also evaluate: (1) The CNN from \cite{deepproblog} that accepts a concatenated set of images as input and is trained to output the final prediction directly. (2) A CNN-LSTM with a Neural Arithmetic Logic Unit (NALU), and (3) a CNN-LSTM with a Neural Accumulator (NAC). For the Hitting Set tasks, we instead use a CNN-LSTM alongside the two \ac{cbm} variants. 

We also perform an ablation study where we firstly assume the symbolic knowledge is given, and secondly, assume the neural network is pre-trained with latent concept labels. \ac{nsl} has equivalent performance in these cases, whilst \textit{learning} symbolic knowledge, and, training \textit{without} latent concept labels. Finally, we demonstrate the effect of the novel corrective examples introduced in this paper, by presenting \ac{nsl} results \textit{without} corrective examples. The ablation results are discussed in more detail in Section \ref{sec:naive_baselines}.

\textbf{Experiment setup.} In the Cumulative Arithmetic tasks, we use the same setup as \cite{ijcai2021-254}, although use ILASP \cite{Law2018thesis} as our symbolic learner. In all other tasks, the neural network in \ac{nsl} is the MNIST CNN from \cite{deepproblog}, and the symbolic learner is FastLAS \cite{law2020fastlas} for the Two-Digit Arithmetic tasks, and ILASP \cite{Law2018thesis} for the Hitting Set tasks. We use ILASP for the Cumulative Arithmetic task as ILASP enables the search space to be expressed using meta-rules. This ensures a fair comparison to $Meta_{Abd}$, which requires a meta-rule specification. ILASP also enables us to learn choice rules in the Hitting Set tasks. We use FastLAS for the Two-Digit Arithmetic tasks to demonstrate NSIL's flexibility to support different symbolic learners, and due to FastLAS' increased efficiency in larger search spaces for this specific class of problems. To evaluate \ac{nsl}, we compare the final prediction to the ground truth label. Each experiment is repeated 20 times using 20 randomly generated seeds, and we measure mean classification accuracy per training epoch, as well as the standard error over the 20 repeats, where 1 epoch = 1 \ac{nsl} iteration. We perform hyper-parameter tuning for all methods using a held-out validation set and a separate random seed. Each experimental repeat is allowed to run for a maximum of 24 hours. The \ac{asp} encodings of the domain knowledge and corrective examples are presented in Appendix A.1.

\subsection{Cumulative Arithmetic}

\begin{table}
\centering
    \resizebox{1\linewidth}{!}{%
    \begin{tabular}{@{}lclllclll@{}}
    \cmidrule(l){2-9}
                    & \multicolumn{4}{c}{\textbf{Addition}}                                                                                        & \multicolumn{4}{c}{\textbf{Product}}                                                                                   \\ \cmidrule(l){2-5} \cmidrule(l){6-9} 
                    & \textbf{Acc.}                                 & \multicolumn{3}{c}{\textbf{MAE}}                                                  & \textbf{Acc.}                                 & \multicolumn{3}{c}{\textbf{log MAE}}                                              \\ \cmidrule(l){2-2} \cmidrule(l){3-5} \cmidrule(l){6-6} \cmidrule(l){7-9}  
    Seq. Len. & 1                                   & \multicolumn{1}{c}{5} & \multicolumn{1}{c}{10} & \multicolumn{1}{c}{100} & 1                                   & \multicolumn{1}{c}{5} & \multicolumn{1}{c}{10} & \multicolumn{1}{c}{15} \\ \midrule
    $Meta_{Abd}$       & \multicolumn{1}{l}{0.953}          & 0.510                & 1.299                 & 6.587                  & \multicolumn{1}{l}{0.977}          & 0.334                & 0.495        & 2.374                 \\ \midrule
    NSIL            & \multicolumn{1}{l}{\textbf{0.984}} & \textbf{0.208}       & \textbf{0.550}        & \textbf{3.960}         & \multicolumn{1}{l}{\textbf{0.983}} & \textbf{0.291}       & \textbf{0.469 }                & \textbf{2.098}        \\ \bottomrule
    \end{tabular}
    }
    \caption{Cumulative Arithmetic task results.}
    \label{tab:recursive}
\end{table}

\begin{figure}
    \centering
    \begin{subfigure}[m]{0.4\linewidth}
        \centering
        \includegraphics[width=\linewidth]{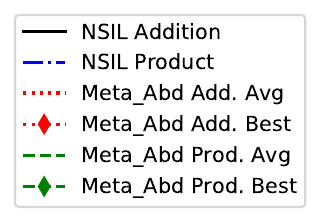}
    \end{subfigure}
    \begin{subfigure}[m]{0.45\linewidth}
        \centering
        \includegraphics[width=1\linewidth]{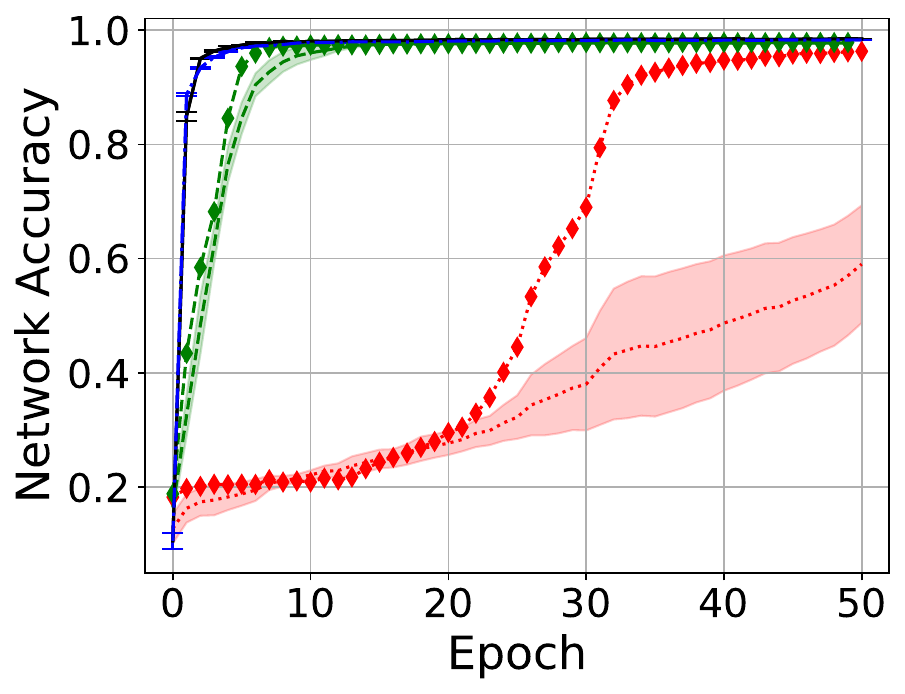}
    \end{subfigure}
    \caption{Cumulative Arithmetic network accuracy.}
    \label{fig:recursive_nn_acc}
\end{figure}

We first consider the Cumulative Addition and Product tasks from \cite{ijcai2021-254}. The dataset $D$ contains $\langle X,y\rangle$ samples where $X$ is a sequence of MNIST digit images, and $y$ is the cumulative addition or product of the digits in $X$. The latent concept has the name $\asp{digit}$ with values $\mathcal{Z} = \{ 0\ ..\ 9\}$. 
The goal is to train the neural network to classify the digit images in $X$, and learn a recursive knowledge base that defines the cumulative arithmetic operation. 
The training set contains 3,000 samples of sequences of length 2-5, and three test sets with sequences of lengths 5, 10, and 100 in the Addition task, and 5, 10, and 15 in the Product task, to verify the trained network and learned knowledge can extrapolate to longer inputs. Each test set contains 10,000 samples. As in \cite{ijcai2021-254}, we evaluate the neural network accuracy (\textbf{Acc.}) on predicting single digits, and the mean average error (\textbf{MAE}), and log mean average error (\textbf{log MAE}) when extrapolating to longer inputs in the addition and product tasks respectively. For this experiment only, we report results over 5 repeats to match \cite{ijcai2021-254}.

In both tasks, \ac{nsl} learns the correct recursive knowledge (see Figure \ref{fig:learned_hyps}), and outperforms $Meta_{Abd}$ (Table \ref{tab:recursive}). \ac{nsl} has a higher neural network accuracy, closer to the state-of-the-art accuracy of 0.9991 when a CNN is trained in a fully supervised fashion \cite{an2020ensemble}, despite \ac{nsl} being trained without latent concept labels. Also, \ac{nsl} trains the neural network to its maximal accuracy with significantly less training epochs than $Meta_{Abd}$ (see Figure \ref{fig:recursive_nn_acc}). $Meta_{Abd}$ relies upon computing the exact psuedo-labels for each image in order to train the network accurately, whereas in \ac{nsl}, NeurASP can train the network with multiple digit choices for each image using a semantic loss function. 
In Figure \ref{fig:recursive_nn_acc}, the error bars and shaded regions indicate standard error over the 5 repeats. We plot the $Meta_{Abd}$ Addition results from their GitHub repository, and re-run their Product task with 5 randomly generated seeds as these results are not available. We also plot the best $Meta_{Abd}$ result compared to the average, to demonstrate that $Meta_{Abd}$ can be highly sensitive to the random seed used to initialise the network in tasks with dense abductive spaces.

\subsection{Two-Digit Arithmetic}\label{sec:eval:arithmetic}

\begin{figure}[t]
    \centering
    \begin{subfigure}[t]{1\linewidth}
            \centering
            \includegraphics[width=1\linewidth]{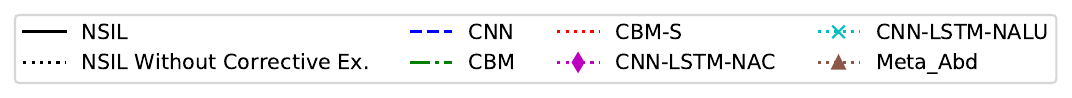}
        \end{subfigure}
        \begin{subfigure}[t]{0.21\textwidth}
            \includegraphics[width=\textwidth]{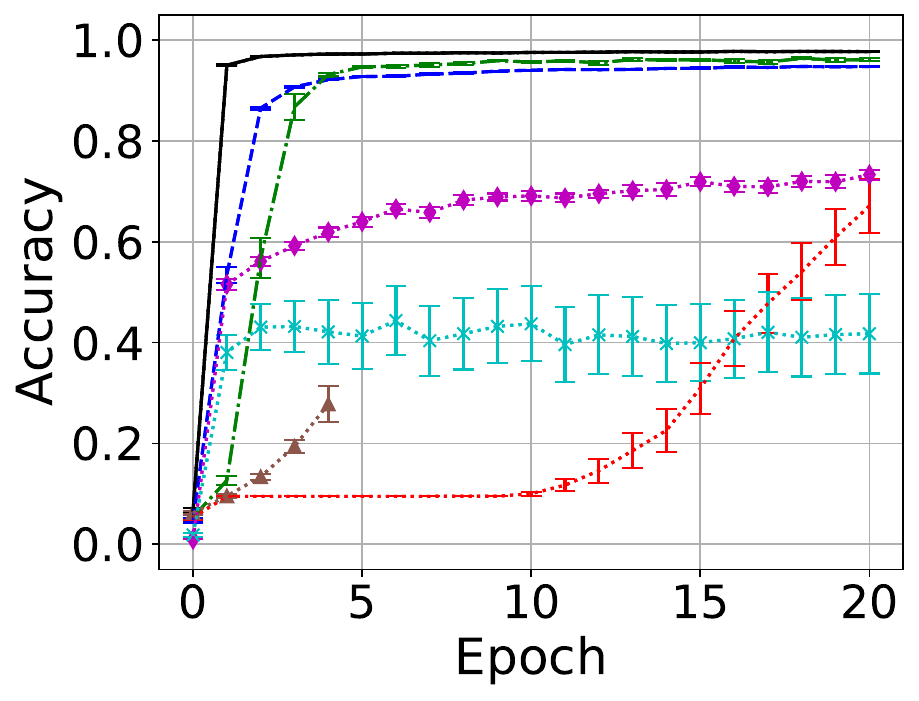}
            \caption{Addition 100\%}
            \label{fig:addition_100_pct}
        \end{subfigure}
        \begin{subfigure}[t]{0.21\textwidth}
            \includegraphics[width=\textwidth]{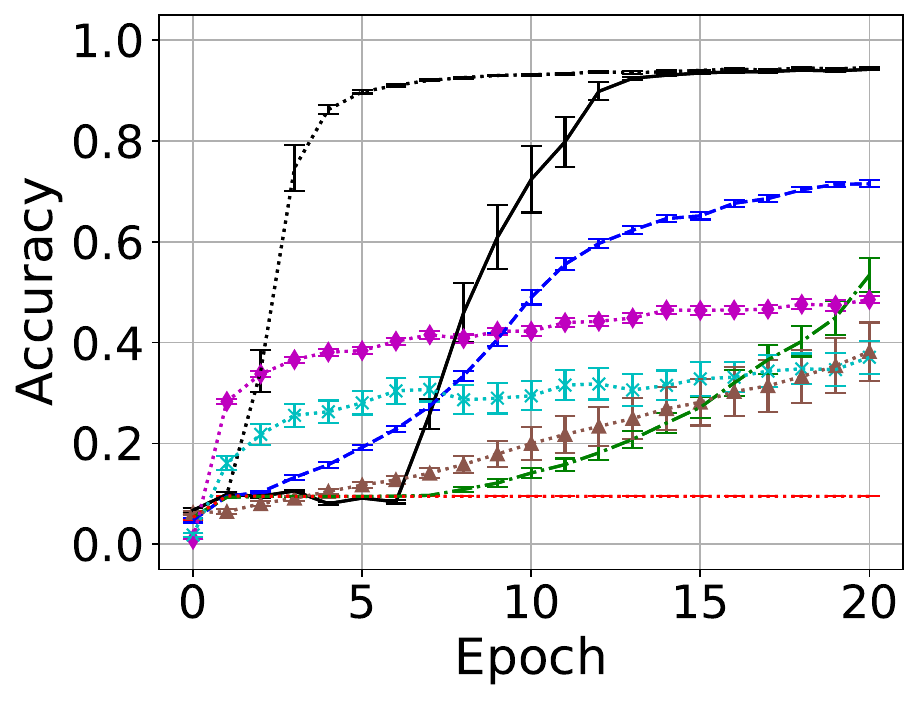}
            \caption{Addition 10\%}
            \label{fig:addition_10_pct}
        \end{subfigure}
        \begin{subfigure}[t]{0.21\textwidth}
            \includegraphics[width=\textwidth]{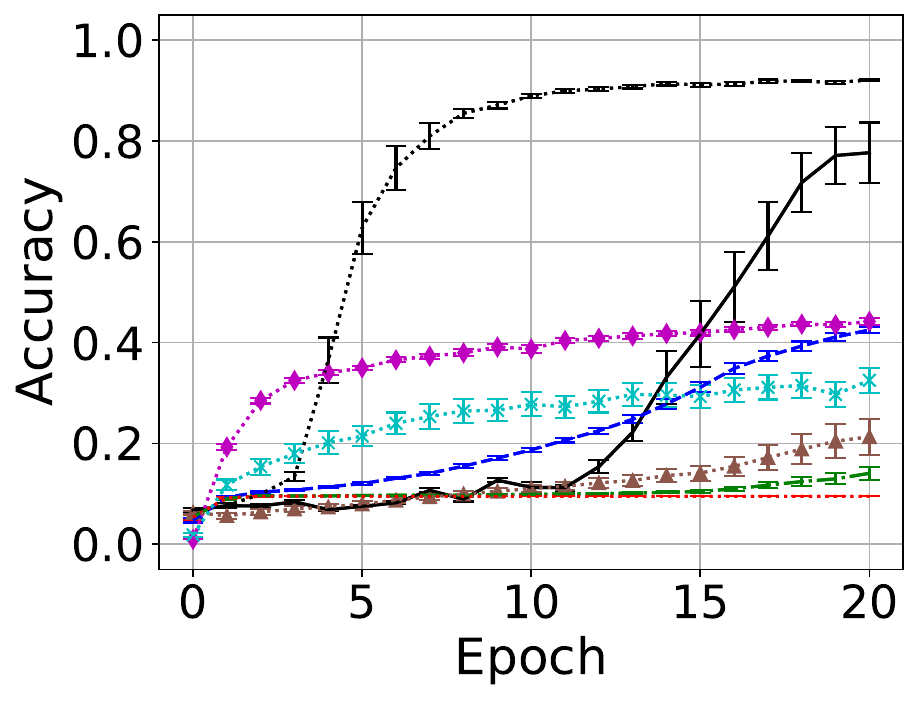}
            \caption{Addition 5\%}
            \label{fig:addition_5_pct}
        \end{subfigure}
        \begin{subfigure}[t]{0.21\textwidth}
            \includegraphics[width=\textwidth]{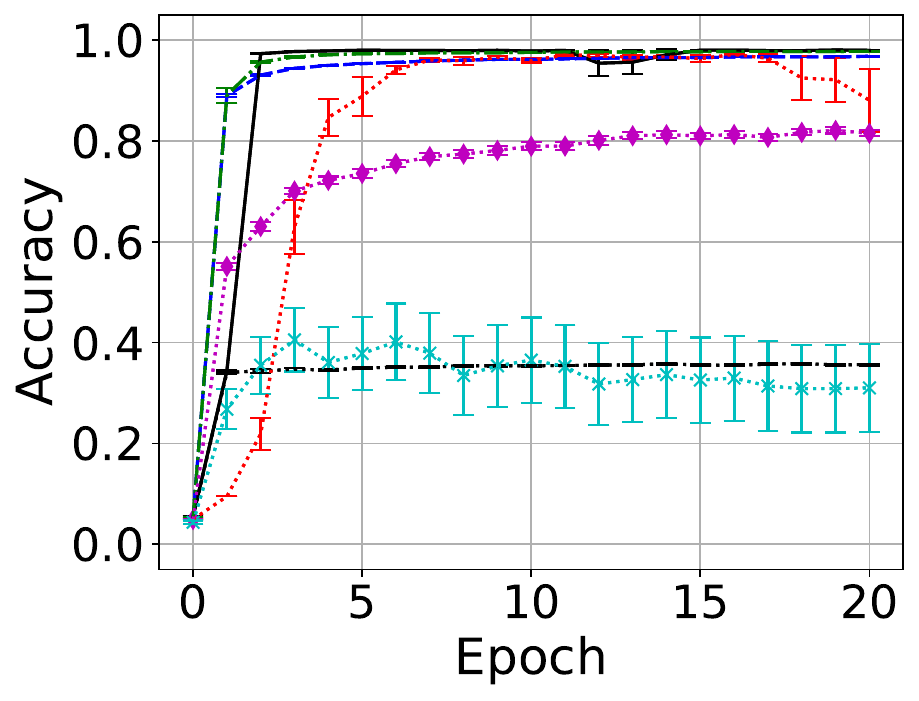}
            \caption{E9P 100\%}
            \label{fig:e9p_100_pct}
        \end{subfigure}
        \begin{subfigure}[t]{0.21\textwidth}
            \includegraphics[width=\textwidth]{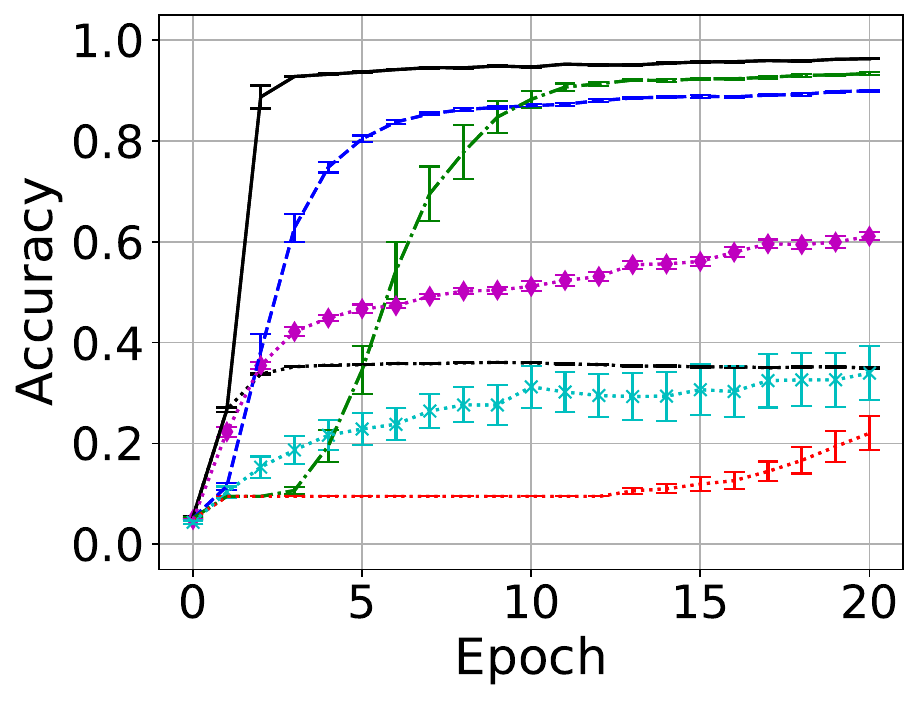}
            \caption{E9P 10\%}
            \label{fig:e9p_10_pct}
        \end{subfigure}
        \begin{subfigure}[t]{0.21\textwidth}
            \includegraphics[width=\textwidth]{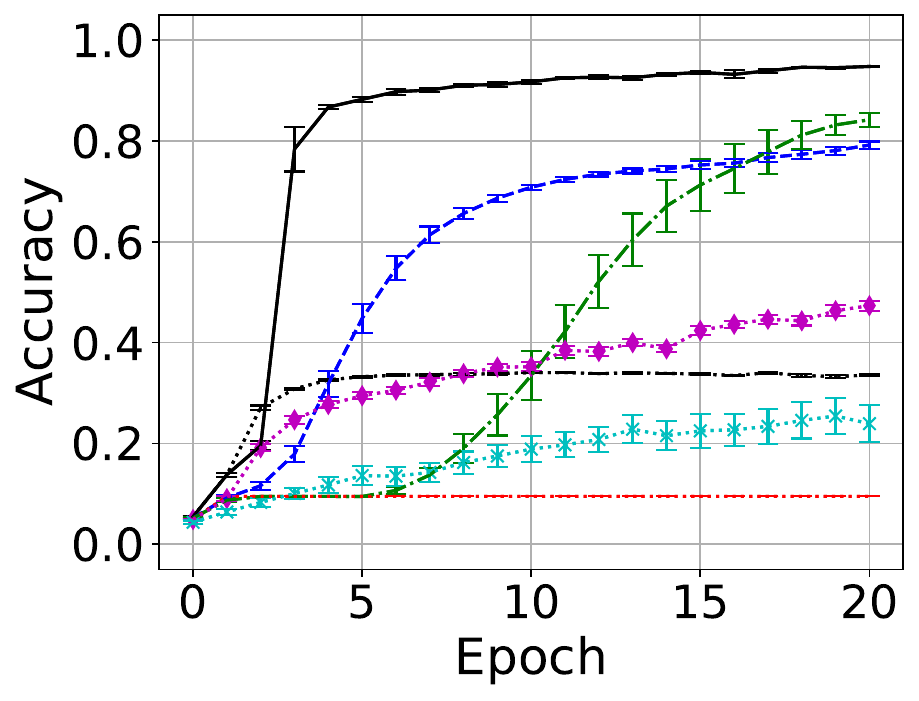}
            \caption{E9P 5\%}
            \label{fig:e9p_5_pct}
        \end{subfigure}
        \caption{Two-Digit Arithmetic results with reduced training sets.}
        \label{fig:mnist_arithmetic_acc}
\end{figure}

\begin{figure}[t]
    \centering
    \begin{subfigure}[t]{0.75\linewidth}
        \centering
        \includegraphics[width=1\linewidth]{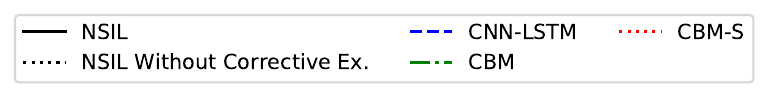}
    \end{subfigure}\vfill
    \begin{subfigure}[b]{0.19\textwidth}
        \centering
        \includegraphics[width=\textwidth]{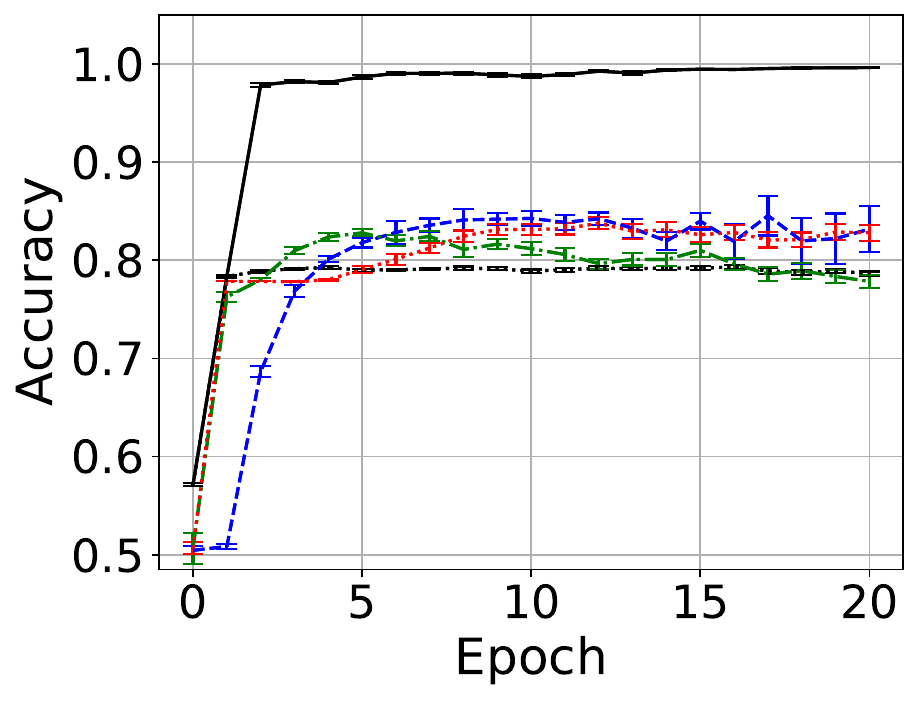}
        \caption{HS MNIST}
    \end{subfigure}
    \begin{subfigure}[b]{0.19\textwidth}
        \centering
        \includegraphics[width=\textwidth]{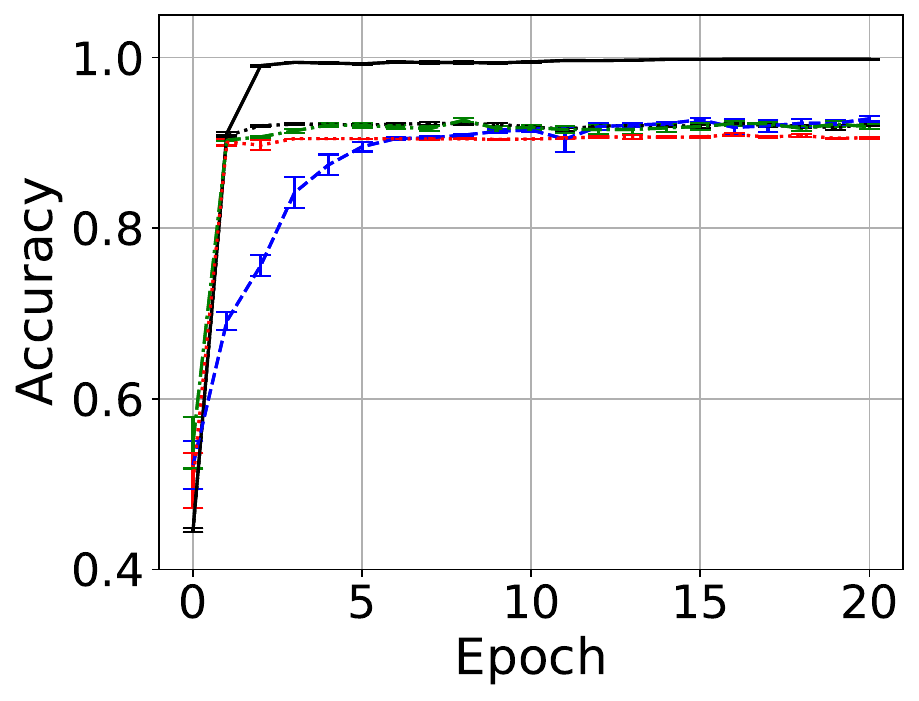}
        \caption{CHS MNIST}
    \end{subfigure}
    \begin{subfigure}[b]{0.19\textwidth}
        \centering
        \includegraphics[width=\textwidth]{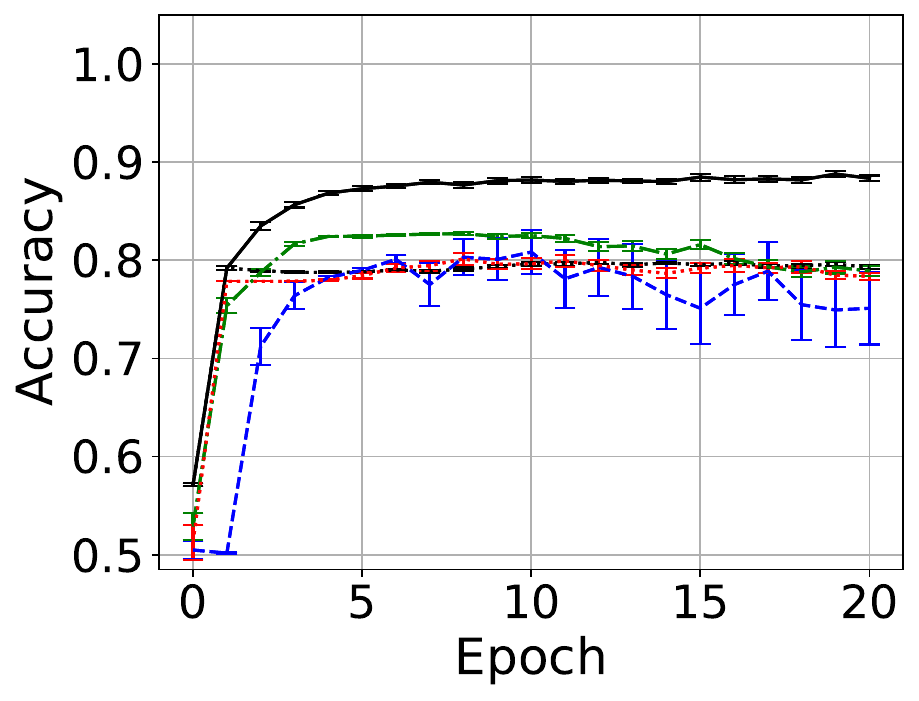}
        \caption{HS FashionMNIST}
    \end{subfigure}
    \begin{subfigure}[b]{0.19\textwidth}
        \centering
        \includegraphics[width=\textwidth]{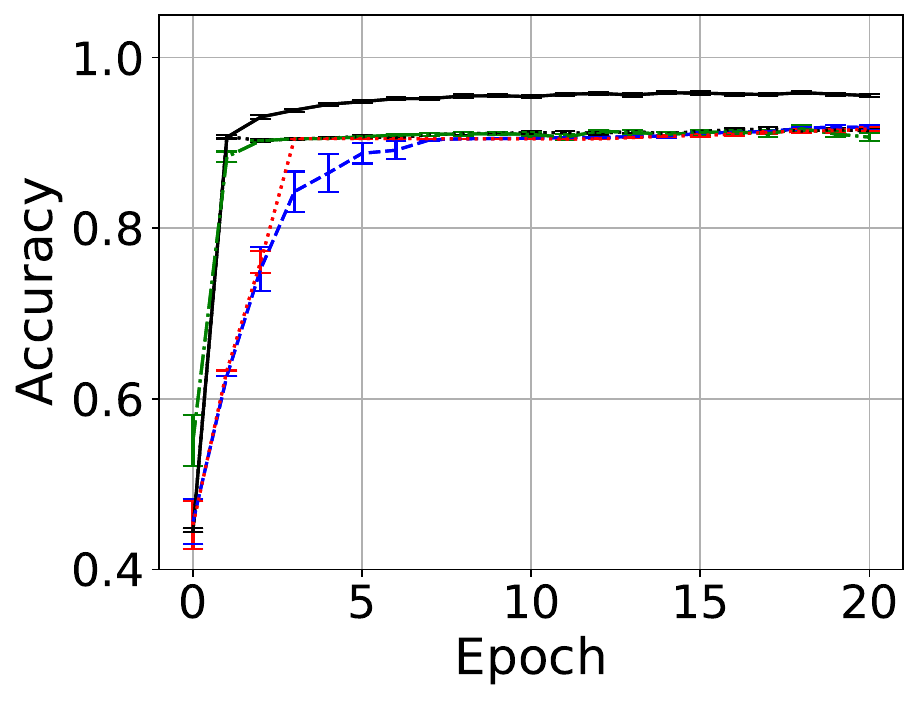}
        \caption{CHS FashionMNIST}
    \end{subfigure}
    
    \caption{Hitting Set accuracy.}
    \label{fig:hitting_sets_accuracy}
\end{figure}

We consider the MNIST Addition task for two single digit numbers used in \cite{deepproblog}, and a variation called MNIST Even9Plus (E9P) that requires learning negation as failure, which is not supported by $Meta_{Abd}$. In this task, the same pair of images are used, but the label is equal to the digit value of the second image, if the digit value of the first image is even, or 9 plus the digit value in the second image otherwise. The purpose of the E9P task is to demonstrate \ac{nsl} can be used to solve multiple tasks within the same domain, i.e., the search space does not have to be engineered \textit{exactly} to solve a particular task. In this case, both sub-tasks are defined over the same set of inputs and outputs, which is more challenging than the Cumulative Arithmetic domain where the output space is significantly different between the Addition and Product sub-tasks. We include a comparison to $Meta_{Abd}$ on the MNIST Addition task, but note that a significantly reduced search space is used compared to \ac{nsl}, as the relations required for E9P are not included, given that $Meta_{Abd}$ does not support learning negation as failure. Both tasks use training, validation and test datasets of size 24,000, 6,000, and 5,000 samples respectively. The relations $\asp{even}$, $\asp{not \ even}$, $\asp{plus\_nine}$, and `=', as well as the function `+' are given in $B$ to specify the search space for the symbolic learner. The latent concept is the same as in the Cumulative Arithmetic tasks.

The overall accuracy results are shown in Figure~\ref{fig:mnist_arithmetic_acc} for reducing percentages of training data (100\% - 5\%). \ac{nsl} outperforms the neural baselines on both tasks for all dataset percentages, and learns the correct knowledge (see Figure \ref{fig:learned_hyps}). On the Addition task, \ac{nsl} achieves superior performance in a much larger search space compared to $Meta_{Abd}$, which also times out after iteration 4 with 100\% of the data. \ac{nsl} trains the neural network to classify MNIST images with a mean accuracy of 0.989, 0.970, and 0.891, for each dataset percentage respectively, which in the 100\% case is within 1\% of the performance of a fully supervised model (0.9991) \cite{an2020ensemble}. For the E9P task, the mean network accuracy is 0.986, 0.974, and 0.962, for each dataset percentage respectively. \ac{nsl} has greater performance in the E9P task than in the Addition task because the label is more informative, despite the same domain knowledge and search space. There is a reduced number of possible latent concept values for certain labels, which gives a stronger back-propagation signal for training the network. 
Finally, we increased the search space in the E9P task with the additional functions; `-', `$\times$', `$\div$', as well as $\asp{plus\_eight},...,\asp{plus\_one}$ relations, and \ac{nsl} converged to the expected knowledge in all of these cases (see Table \ref{tab:mnist_e9p_increasing_hyp_space}).

\begin{figure*}[t]
    \centering
    \begin{subfigure}[m]{0.25\textwidth}
        \begin{lstlisting}
<--Addition--!>
f(A,B) :- eq(A,B).
f(A,B) :- add(A,C), f(C,B).

<--Product--!>
f(A,B) :- eq(A,B).
f(A,B) :- mult(A,C), f(C,B).
    \end{lstlisting}
    \caption{Cumulative Arithmetic}
    \label{fig:recursive_arithmetic_learned_hyps}
    \end{subfigure}
    \hspace{10pt}
    \begin{subfigure}[m]{0.31\textwidth}
        \begin{lstlisting}
<--Addition--!>
result(V0,V1,V2) :- V2 = V0 + V1.

<--E9P--!>
result(V0,V1,V2) :- even(V0),V2 = V1. 
result(V0,V1,V2) :- not even(V0), plus_nine(V1,V2).
    \end{lstlisting}    
        \caption{Two-Digit Arithmetic}
        \label{fig:mnist_arithmetic_learned_hyps}
    \end{subfigure}
    \hspace{10pt}
    \begin{subfigure}[m]{0.34\textwidth}
        \begin{lstlisting}
<--HS--!>
0 {hs(V1,V2) } 1.
hit(V1) :- hs(V3,V2), ss_element(V1,V2).
:- ss_element(V1,V2), not hit(V1).
:- hs(V3,V1), hs(V3,V2), V1 != V2.
<--CHS--!>
0 {hs(V1,V2) } 1.
hit(V1) :- hs(V3,V2), ss_element(V1,V2).
:- ss_element(V1,V2), not hit(V1).
:- hs(V3,V1), hs(V3,V2), V1 != V2.
:- ss_element(3,V2), ss_element(V1,1).
    \end{lstlisting}
        \caption{Hitting Sets}
        \label{fig:hitting_sets_learned_hyps}
        \end{subfigure}
        \caption{\ac{nsl} learned knowledge.}
        \label{fig:learned_hyps}
\end{figure*}

\subsection{Hitting Set Problem}\label{sec:eval:hitting_sets}
We consider two variations of the Hitting Set problem \cite{karp1972reducibility}. The first is the standard formulation: Consider a universe of elements $U$, and a collection of sets $S$ which contains subsets of $U$. A hitting set is a special set $ \Phi \subseteq U$ that intersects with every subset in $S$. Given $S$ and an integer $k$, the task is to decide if there is a hitting set of $S$ with size $\leq k$.

\begin{example}
Let $U=\{1\ ..\ 5\}$, $S_{1}=\{\{1\},\{1,3\},\{4\}\}$, and $k=2$. $S_{1}$ has a hitting set of size $\leq 2$: $\Phi=\{1, 4\}$. We say that each subset in $S$ is ``hit" if it intersects with the hitting set. Now consider $S_{2}=\{\{1\},\{2,3\},\{4\}\}$. $S_{2}$ has no hitting set of size $\leq 2$. 
\end{example}

In our experiments, we replace elements $\{1\ ..\ 5\}$ in $U$ with corresponding MNIST images from classes 1-5 (i.e., $\mathcal{Z}=\{1 \ ..\ 5 \}$). We assume $k=2$ and construct $\langle X,y\rangle$ data points where $X$ represents a collection of sets $S$, and $y$ is a binary label indicating whether $X$ has a hitting set of size $\leq k$. We also assume a maximum of 4 elements, $X$ is well-formed, with no duplicate elements in a set, and elements within the same set are arranged in ascending order. For example; $X= \left \{ \{ \inlinegraphics{figures/digits/1.png} \}, \{\inlinegraphics{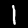}, \inlinegraphics{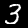} \}, \{ \inlinegraphics{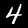}\} \right \}$, $y=1$. We denote this standard hitting set variant as HS. To demonstrate \ac{nsl} can solve multiple tasks within the same domain, we introduce a second variant, denoted CHS, which adds the constraint that no hitting sets occur if $\vert X \vert \geq 3$, and element 1 (in image form) exists in any subset in $X$. We also present results for both variants with a more difficult perception task using FashionMNIST images instead of MNIST. The goal is to train the neural network to classify images, whilst learning the HS or CHS rules to decide the existence of a hitting set. Training, validation and test datasets contain 1502, 376, and 316 examples respectively. Each latent concept value is represented in \ac{asp} as $\asp{ss\_element(}s,z)$, where $s\in \{1\ ..\ 4\}$ is a subset identifier, and $z \in \{1\ ..\ 5\}$ is the associated element. The domain knowledge $B$ contains $k=2$, the relation `!=' and the relation $\asp{hs}$, which defines an element of the hitting set. $B$ also contains the relations for the CHS task: $\asp{ss\_element(3,V_{1})}$ and $\asp{ss\_element(V_{2},1)}$ that define respectively subset 3 (and any element $\asp{V_{1}}$), and element $1$ in any subset $\asp{V_{2}}$. In these tasks, we use a CNN-LSTM baseline in addition to the \ac{cbm} variants, where a CNN firstly encodes a feature vector for each image in a sequence $X$, and an LSTM returns a binary classification as to the existence of a hitting set. To encode the subset structure of $X$ for the CNN-LSTM, we create an image of `\{' and `\}' characters and pass in the entire sequence as input. Note that in \ac{nsl}, the structure is passed directly into the symbolic learner using the $\asp{ss\_element}$ relations.

Figure~\ref{fig:hitting_sets_accuracy} shows the overall accuracy for both Hitting Set tasks using MNIST and FashionMNIST images. \ac{nsl} outperforms the baselines on both tasks and learns the correct knowledge (see Figure \ref{fig:learned_hyps}). When classifying images, the neural network achieves a mean accuracy of 0.993 (MNIST) and 0.897 (FashionMNIST) on the HS task, and 0.993 (MNIST) and 0.894 (FashionMNIST) on the CHS task. Note that the neural network accuracy is for concepts 1-5 only. All methods perform better on the CHS task than the HS task. This is due to a shortcut; a negative test example with 3 subsets can be correctly predicted if the neural network predicts element 1 for \textit{any} image within the subset, as opposed to having to rely on accurate predictions for more images in the HS task. 

The knowledge learned by \ac{nsl} has multiple answer sets as there could be multiple hitting sets of a given collection. Therefore, despite training labels indicating the existence of a hitting set, \ac{nsl} can generate \textit{all} the hitting sets at inference time. We verified this by computing the hamming loss against the ground truth hitting sets for all test samples. The results are 0.003 and 0.150 on the HS task, and 0.003 and 0.126 on the CHS task, for the MNIST and FashionMNIST images respectively. This indicates almost perfect hitting set generation, with the minor errors due to mis-classifications by the neural network, as opposed to errors in the learned knowledge. Also, it is crucial to note that the learned knowledge does not depend on the number of elements, the value of $k$, or the possible latent concept values, despite these restrictions during training. This demonstrates \ac{nsl} is able to \textit{generalise} beyond the training setup. Finally, in the more challenging HS FashionMNIST task, we also sampled 6 sets of additional $\asp{ss\_element}$ domain relations to extend the search space for the symbolic learner. The results are presented in Table~\ref{tab:hs_increasing_hyp_space}, and \ac{nsl} learned the correct knowledge in all of these cases, but sometimes required more iterations and/or time to do so. The \textit{Standard} domain knowledge contains only the predicates required to learn the expected rules of the HS task. This includes $k=2$, the `!=' relation, and the relation $\asp{hs}$ which defines an element of the hitting set. The additional $\asp{ss\_element}$ relations contain a combination of elements $\{1\ ..\ 5\}$ (denoted `el') and subset identifiers $\{1\ ..\ 4\}$ (denoted `ssID').

\begin{table}[H]
    \begin{minipage}{1\linewidth}
      \centering
        \resizebox{0.87\linewidth}{!}{%
        \begin{tabular}{ccccc}
        \toprule
         & \textbf{\makecell{Domain\\ knowledge}} & \textbf{\makecell{Convergence\\iteration}} & \textbf{\makecell{Convergence\\time (s)}} \\
        \midrule
        * & \makecell{+, =, $\asp{even}$, $\asp{not\ even}$,\\$\asp{plus\_nine}$} & 2 & 475.44 \\
        \midrule
        & \makecell{+, -, $\times$, $\div$, =, $\asp{even}$,\\ $\asp{not\ even}$, $\asp{plus\_nine}$} &                                  2 &                         547.81 \\
        \midrule
        & \makecell{=, $\asp{even}$, $\asp{not\ even}$,\\ $\asp{plus\_nine},...,\asp{plus\_one}$} &                                  1 &                        \textit{402.03} \\
        \midrule
        & \makecell{+, =, $\asp{even}$, $\asp{not\ even}$,\\ $\asp{plus\_nine},...,\asp{plus\_one}$} &                                  2 &                        \textit{881.35} \\
        \midrule
        & \makecell{+, -, $\times$, $\div$, =, $\asp{even}$, $\asp{not\ even}$,\\ $\asp{plus\_nine},...,\asp{plus\_one}$} &                                  2 &                        \textit{57869.12} \\
        
        \bottomrule
        \end{tabular}%
        }
        \subcaption{E9P}
        \label{tab:mnist_e9p_increasing_hyp_space}
    \end{minipage}%
    \vspace{1em}
    \begin{minipage}{1\linewidth}
      \centering
        \resizebox{0.9\linewidth}{!}{%
        \begin{tabular}{cccc}
        \toprule
        & \textbf{\makecell{Domain\\knowledge}} & \textbf{\makecell{Convergence\\iteration}} & \textbf{\makecell{Convergence\\time (s)}} \\
        \midrule
        & Standard, ssID 2, ssID 4 & 1 & 111.34 \\ \midrule & Standard, ssID 2, el 2 & 1 & 113.21 \\ \midrule & Standard & 1 & 115.77 \\ \midrule & Standard, ssID 1, el 4 & 1 & 130.71 \\ \midrule & Standard, ssID 3, el 4 & 2 & 734.70 \\ \midrule* & Standard, ssID 3, el 1 & 2 & 852.84 \\ \midrule & Standard, ssID 1, ssID 3 & 2 & 888.54 \\
        \bottomrule
        \end{tabular}
        }
        \subcaption{HS FashionMNIST}
        \label{tab:hs_increasing_hyp_space}
    \end{minipage} 
    \caption{Increasing the search space for the symbolic learner. The asterisks indicate the configuration used in the main results. Run-times in italics indicate that weight pruning is used to remove corrective examples with weight $<$ 5, to obtain results efficiently.}
    \vspace{-4pt}
\end{table}

\begin{table*}[t]
\centering
\resizebox{0.8\linewidth}{!}{%
\begin{tabular}{@{}lllllll@{}}
\cmidrule(l){2-7}
\multicolumn{1}{c}{} & \multicolumn{3}{c}{\textbf{Addition}}                                             & \multicolumn{3}{c}{\textbf{E9P}}                                                  \\ \cmidrule(l){2-7}
Dataset \% & \multicolumn{1}{c}{100} & \multicolumn{1}{c}{10} & \multicolumn{1}{c}{5} & \multicolumn{1}{c}{100} & \multicolumn{1}{c}{10} & \multicolumn{1}{c}{5} \\ \midrule
FF-NSL & \textbf{0.9753} (0.0021) & 0.9362 (0.0029) & \textbf{0.9151} (0.0058) & \textbf{0.9809} (0.0016) & 0.9513 (0.0030) & 0.9346 (0.0051) \\
NeurASP & \textbf{0.9762} (0.0013) & \textbf{0.9492} (0.0016) & \textbf{0.9149} (0.0051) & \textbf{0.9797} (0.0015) & \textbf{0.9642} (0.0009) & \textbf{0.9500} (0.0014) \\ \midrule
NSIL & \textbf{0.9762} (0.0013) & 0.9449 (0.0025) & 0.8782 (0.0134) & \textbf{0.9816} (0.0009) & \textbf{0.9634} (0.0007) & \textbf{0.9510} (0.0016) \\ \bottomrule
\end{tabular}
}
\caption{Two-Digit Arithmetic ablation results. Best results highlighted in bold (we declare a tie in the case of overlapping error bars).}
\label{tab:non_recursive_naive}
\end{table*}

\begin{table*}[t]
\centering
\resizebox{0.58\linewidth}{!}{%
\begin{tabular}{@{}lllll@{}}
\cmidrule(l){2-5}
\multicolumn{1}{c}{}        & \multicolumn{2}{c}{\textbf{HS}}                                       & \multicolumn{2}{c}{\textbf{CHS}}                                      \\ \cmidrule(l){2-5} 
Images & \multicolumn{1}{c}{MNIST} & \multicolumn{1}{c}{FashionMNIST} & \multicolumn{1}{c}{MNIST} & \multicolumn{1}{c}{FashionMNIST} \\ \midrule
FF-NSL & 0.9937 (0.0017) & 0.8816 (0.0110) & 0.9962 (0.0012) & \textbf{0.9563} (0.0034) \\
NeurASP & \textbf{0.9981} (0.0013) & \textbf{0.8975} (0.0041) & \textbf{0.9994} (0.0006) & \textbf{0.9538} (0.0070) \\ \midrule
NSIL & \textbf{0.9962} (0.0012) & 0.8747 (0.0053) & \textbf{0.9981} (0.0013) & \textbf{0.9544} (0.0021) \\ \bottomrule
\end{tabular}
}
\caption{Hitting Set ablation results. Best results highlighted in bold (we declare a tie in the case of overlapping error bars).}
\label{tab:hitting_sets_naive}
\end{table*}

\section{Ablations}\label{sec:naive_baselines}
In this section, we present three ablations: (1) We remove the requirement to learn the symbolic knowledge and assume the knowledge is given. (2) We pre-train the neural network with manually annotated latent concept labels, and (3) we evaluate \ac{nsl} without corrective examples in order to demonstrate the impact of the novel contribution of this paper. To implement the first two ablations, we respectively evaluate NeurASP \cite{neurasp}, which assumes the symbolic knowledge is given, and FF-NSL \cite{cunnington2023ffnsl} which pre-trains the neural network on a separate dataset. FF-NSL then uses the neural network predictions to train the symbolic learner. We provide a comparison to \ac{nsl}, and train each method on the experiments presented in Section \ref{sec:experiments} for 20 epochs (where 1 epoch = 1 \ac{nsl} iteration). We evaluate the mean performance and standard error over 5 repeats.

The Two-Digit Arithmetic results are presented in Table \ref{tab:non_recursive_naive} for reducing percentages of training data (100\%\ -\ 5\%). The results indicate final task accuracy after performing a forward pass on the neural and symbolic components. In both the Addition and E9P tasks, \ac{nsl} achieves comparable accuracy to the fully supervised approach of FF-NSL, even when \textit{no latent concept labels} are observed during training. Also, \ac{nsl} achieves comparable accuracy to NeurASP, whilst also \textit{learning} the symbolic knowledge. The Hitting Set results are shown in Table \ref{tab:hitting_sets_naive}, where the results also indicate final task accuracy. Again, \ac{nsl} performs comparably to FF-NSL and NeurASP in each experiment.

In the final ablation study, the corrective examples enable \ac{nsl} to converge to a more accurate solution in most cases, by biasing the symbolic learner to explore the symbolic search space. This is evidenced in the E9P task shown in Figures \ref{fig:e9p_100_pct}-\ref{fig:e9p_5_pct}, where the correct knowledge is \textit{not} learned on the first iteration (compare \textit{\ac{nsl}} to \textit{\ac{nsl} Without Corrective Ex.}). In cases where the correct knowledge \textit{is} learned on the first iteration, the effect of the corrective examples depends on the size of the dataset. When there are sufficient data points to train the neural network accurately, the corrective examples don't offer any improvement, because the network can be trained given the correct knowledge (see Figure \ref{fig:addition_100_pct}). For small datasets, the corrective examples together with a sub-optimal network may lead to a delay in convergence compared to the case where there are no corrective examples (see Figures \ref{fig:addition_10_pct} and \ref{fig:addition_5_pct}). However, in real-world applications, we consider these to be very special cases, as it's unlikely the correct knowledge will be learned on the first iteration. Therefore, corrective examples are important to \ac{nsl}'s performance.

\section{Conclusion}
This paper presents \ac{nsl} which learns expressive knowledge to solve computationally complex problems, whilst training a general neural network to classify latent concepts from raw data. The novelty is the method of biasing the symbolic learner using corrective examples weighted to balance exploration and exploitation of the symbolic search space. The results show that \ac{nsl} outperforms the closest existing system $Meta_{Abd}$, and a variety of purely neural baselines, whilst learning more expressive and general knowledge.

\textbf{Limitations}. During NSIL's iterative training cycle, using the corrective examples may require more iterations to converge given a small dataset. However, the benefit is that a high accuracy is reached in cases where the correct knowledge is not learned on the first iteration. Also, the neural baselines learn faster than \ac{nsl}, although given more time, \ac{nsl} achieves a greater accuracy (see Appendix A.2).

\textbf{Future work}. One could investigate theoretical properties of \ac{nsl}'s corrective examples, and also apply \ac{nsl} to more challenging tasks with more complex imagery. In these cases, the neural network could be trained for more epochs on each \ac{nsl} iteration, or utilise pre-trained weights on a larger dataset such as ImageNet. This would enable \ac{nsl} to be used as a ``fine-tuning" approach to learn the weights of a linear layer for classifying latent concepts. Finally, future work could generalise \ac{nsl} to support multiple latent concepts by integrating multiple neural networks, and extend \ac{nsl} to take advantage of unsupervised learning techniques.

\section*{Ethical Statement}
There are no direct negative impacts that we can envisage for our work, given we introduce a general machine learning approach. However, \ac{nsl} inherits general concerns regarding the deployment of machine learning systems, and appropriate precautions should be taken, such as ensuring training data is unbiased, and model inputs/outputs are monitored for adversarial attacks. As \ac{nsl} learns human interpretable knowledge using symbolic learning, this may help to mitigate these issues in some applications, by revealing potential bias, and providing a level of assurance regarding possible downstream predictions based on the learned knowledge. The usual performance monitoring will still be required if \ac{nsl} is deployed into production, to prevent adversarial attacks, and to detect when re-training may be required if the input data is subject to distributional shifts.

\section*{Acknowledgements}
The authors would like to thank Krysia Broda for helpful discussions, and also the reviewers for their insightful comments.

\bibliographystyle{named}
\bibliography{ijcai23}

\begin{thebibliography}{}

\bibitem[\protect\citeauthoryear{An \bgroup \em et al.\egroup
  }{2020}]{an2020ensemble}
Sanghyeon An, Minjun Lee, Sanglee Park, Heerin Yang, and Jungmin So.
\newblock An ensemble of simple convolutional neural network models for mnist
  digit recognition.
\newblock {\em arXiv preprint arXiv:2008.10400}, 2020.

\bibitem[\protect\citeauthoryear{Aspis \bgroup \em et al.\egroup
  }{2022}]{aspis2022embed2sym}
Yaniv Aspis, Krysia Broda, Jorge Lobo, and Alessandra Russo.
\newblock Embed2sym-scalable neuro-symbolic reasoning via clustered embeddings.
\newblock In {\em Proceedings of the International Conference on Principles of
  Knowledge Representation and Reasoning}, volume~19, pages 421--431, 2022.

\bibitem[\protect\citeauthoryear{Badreddine \bgroup \em et al.\egroup
  }{2022}]{Badreddine2022}
Samy Badreddine, Artur {d'Avila Garcez}, Luciano Serafini, and Michael
  Spranger.
\newblock Logic tensor networks.
\newblock {\em Artificial Intelligence}, 303:103649, 2022.

\bibitem[\protect\citeauthoryear{Besold \bgroup \em et al.\egroup
  }{2022}]{Besold2017}
Tarek~R Besold, Artur~d’Avila Garcez, Sebastian Bader, Howard Bowman, Luis~C
  Lamb, Leo de~Penning, BV~Illuminoo, Hoifung Poon, and COPPE Gerson~Zaverucha.
\newblock Neural-symbolic learning and reasoning: A survey and interpretation.
\newblock {\em Neuro-Symbolic Artificial Intelligence: The State of the Art},
  342:1, 2022.

\bibitem[\protect\citeauthoryear{Corapi \bgroup \em et al.\egroup
  }{2010}]{Corapi10}
Domenico Corapi, Alessandra Russo, and Emil Lupu.
\newblock Inductive logic programming as abductive search.
\newblock In {\em Technical communications of the 26th international conference
  on logic programming}, pages 54--63. Schloss Dagstuhl-Leibniz-Zentrum fuer
  Informatik, 2010.

\bibitem[\protect\citeauthoryear{Cunnington \bgroup \em et al.\egroup
  }{2023}]{cunnington2023ffnsl}
Daniel Cunnington, Mark Law, Jorge Lobo, and Alessandra Russo.
\newblock Ffnsl: Feed-forward neural-symbolic learner.
\newblock {\em Machine Learning}, pages 1--55, 2023.

\bibitem[\protect\citeauthoryear{Dai and Muggleton}{2021}]{ijcai2021-254}
Wang-Zhou Dai and Stephen Muggleton.
\newblock Abductive knowledge induction from raw data.
\newblock In Zhi-Hua Zhou, editor, {\em Proceedings of the Thirtieth
  International Joint Conference on Artificial Intelligence, {IJCAI-21}}, pages
  1845--1851. International Joint Conferences on Artificial Intelligence
  Organization, 8 2021.
\newblock Main Track.

\bibitem[\protect\citeauthoryear{Dai \bgroup \em et al.\egroup
  }{2019}]{dai2019bridging}
Wang-Zhou Dai, Qiuling Xu, Yang Yu, and Zhi-Hua Zhou.
\newblock Bridging machine learning and logical reasoning by abductive
  learning.
\newblock {\em Advances in Neural Information Processing Systems}, 32, 2019.

\bibitem[\protect\citeauthoryear{Daniele \bgroup \em et al.\egroup
  }{2022}]{dsl}
Alessandro Daniele, Tommaso Campari, Sagar Malhotra, and Luciano Serafini.
\newblock Deep symbolic learning: Discovering symbols and rules from
  perceptions.
\newblock {\em arXiv preprint arXiv:2208.11561}, 2022.

\bibitem[\protect\citeauthoryear{Dantsin \bgroup \em et al.\egroup
  }{2001}]{dantsin2001complexity}
Evgeny Dantsin, Thomas Eiter, Georg Gottlob, and Andrei Voronkov.
\newblock Complexity and expressive power of logic programming.
\newblock {\em ACM Computing Surveys (CSUR)}, 33(3):374--425, 2001.

\bibitem[\protect\citeauthoryear{Dasaratha \bgroup \em et al.\egroup
  }{2023}]{Deeppsl}
Sridhar Dasaratha, Sai~Akhil Puranam, Karmvir~Singh Phogat, Sunil~Reddy
  Tiyyagura, and Nigel~P. Duffy.
\newblock Deeppsl: End-to-end perception and reasoning, 2023.

\bibitem[\protect\citeauthoryear{d'Avila Garcez \bgroup \em et al.\egroup
  }{2019}]{GarcezGLSST19}
Artur d'Avila Garcez, Marco Gori, Luís~C. Lamb, Luciano Serafini, Michael
  Spranger, and Son~N. Tran.
\newblock Neural-symbolic computing: An effective methodology for principled
  integration of machine learning and reasoning.
\newblock {\em FLAP}, 6(4):611--632, 2019.

\bibitem[\protect\citeauthoryear{De~Raedt \bgroup \em et al.\egroup
  }{2020}]{ijcai2020-0688}
Luc De~Raedt, Sebastijan Dumančić, Robin Manhaeve, and Giuseppe Marra.
\newblock From statistical relational to neuro-symbolic artificial
  intelligence.
\newblock In Christian Bessiere, editor, {\em Proceedings of the Twenty-Ninth
  International Joint Conference on Artificial Intelligence, {IJCAI-20}}, pages
  4943--4950. International Joint Conferences on Artificial Intelligence
  Organization, 7 2020.
\newblock Survey track.

\bibitem[\protect\citeauthoryear{Evans and
  Grefenstette}{2018}]{evans2018learning}
Richard Evans and Edward Grefenstette.
\newblock Learning explanatory rules from noisy data.
\newblock {\em Journal of Artificial Intelligence Research}, 61:1--64, 2018.

\bibitem[\protect\citeauthoryear{Evans \bgroup \em et al.\egroup
  }{2021}]{EVANS2021103521}
Richard Evans, Matko Bošnjak, Lars Buesing, Kevin Ellis, David Pfau, Pushmeet
  Kohli, and Marek Sergot.
\newblock Making sense of raw input.
\newblock {\em Artificial Intelligence}, 299:103521, 2021.

\bibitem[\protect\citeauthoryear{Ferreira \bgroup \em et al.\egroup
  }{2022}]{ferreira2022looking}
Joao Ferreira, Manuel de~Sousa~Ribeiro, Ricardo Gon{\c{c}}alves, and Joao
  Leite.
\newblock Looking inside the black-box: Logic-based explanations for neural
  networks.
\newblock In {\em Proceedings of the International Conference on Principles of
  Knowledge Representation and Reasoning}, volume~19, pages 432--442, 2022.

\bibitem[\protect\citeauthoryear{Gelfond and Kahl}{2014}]{gelfond2014knowledge}
Michael Gelfond and Yulia Kahl.
\newblock {\em Knowledge representation, reasoning, and the design of
  intelligent agents: The answer-set programming approach}.
\newblock Cambridge University Press, Cambridge, UK, 2014.

\bibitem[\protect\citeauthoryear{Han \bgroup \em et al.\egroup
  }{2021}]{han2021unifying}
Zhongyi Han, Benzheng Wei, Xiaoming Xi, Bo~Chen, Yilong Yin, and Shuo Li.
\newblock Unifying neural learning and symbolic reasoning for spinal medical
  report generation.
\newblock {\em Medical Image Analysis}, 67:101872, 2021.

\bibitem[\protect\citeauthoryear{Karp}{1972}]{karp1972reducibility}
Richard~M Karp.
\newblock Reducibility among combinatorial problems.
\newblock In {\em Complexity of computer computations}, pages 85--103.
  Springer, 1972.

\bibitem[\protect\citeauthoryear{Koh \bgroup \em et al.\egroup
  }{2020}]{koh2020concept}
Pang~Wei Koh, Thao Nguyen, Yew~Siang Tang, Stephen Mussmann, Emma Pierson, Been
  Kim, and Percy Liang.
\newblock Concept bottleneck models.
\newblock In {\em International Conference on Machine Learning}, pages
  5338--5348, 2020.

\bibitem[\protect\citeauthoryear{Law \bgroup \em et al.\egroup
  }{2019}]{LawRB19}
Mark Law, Alessandra Russo, and Krysia Broda.
\newblock Logic-based learning of answer set programs.
\newblock In {\em Reasoning Web. Explainable Artificial Intelligence - 15th
  International Summer School 2019, Bolzano, Italy, September 20-24, 2019,
  Tutorial Lectures}, pages 196--231, 2019.

\bibitem[\protect\citeauthoryear{Law \bgroup \em et al.\egroup
  }{2020}]{law2020fastlas}
Mark Law, Alessandra Russo, Elisa Bertino, Krysia Broda, and Jorge Lobo.
\newblock Fastlas: scalable inductive logic programming incorporating
  domain-specific optimisation criteria.
\newblock In {\em Proceedings of the AAAI Conference on Artificial
  Intelligence}, volume~34, pages 2877--2885, 2020.

\bibitem[\protect\citeauthoryear{Law}{2018}]{Law2018thesis}
Mark Law.
\newblock {\em Inductive learning of answer set programs}.
\newblock PhD thesis, Imperial College London, 2018.

\bibitem[\protect\citeauthoryear{Lyn~Paul \bgroup \em et al.\egroup
  }{2019}]{lyn2019opportunities}
Celeste Lyn~Paul, Leslie~M Blaha, Corey~K Fallon, Cleotilde Gonzalez, and
  Robert~S Gutzwiller.
\newblock Opportunities and challenges for human-machine teaming in
  cybersecurity operations.
\newblock In {\em Proceedings of the human factors and ergonomics society
  annual meeting}, volume~63, pages 442--446. SAGE Publications Sage CA: Los
  Angeles, CA, 2019.

\bibitem[\protect\citeauthoryear{Manhaeve \bgroup \em et al.\egroup
  }{2018}]{deepproblog}
Robin Manhaeve, Sebastijan Dumancic, Angelika Kimmig, Thomas Demeester, and Luc
  De~Raedt.
\newblock Deepproblog: Neural probabilistic logic programming.
\newblock In {\em Advances in Neural Information Processing Systems}, pages
  3749--3759, 2018.

\bibitem[\protect\citeauthoryear{Muggleton and
  De~Raedt}{1994}]{muggleton1994inductive}
Stephen Muggleton and Luc De~Raedt.
\newblock Inductive logic programming: Theory and methods.
\newblock {\em The Journal of Logic Programming}, 19:629--679, 1994.

\bibitem[\protect\citeauthoryear{Muggleton \bgroup \em et al.\egroup
  }{2015}]{Muggleton13}
Stephen~H Muggleton, Dianhuan Lin, and Alireza Tamaddoni-Nezhad.
\newblock Meta-interpretive learning of higher-order dyadic datalog: Predicate
  invention revisited.
\newblock {\em Machine Learning}, 100(1):49--73, 2015.

\bibitem[\protect\citeauthoryear{Payani and Fekri}{2019}]{payani2019inductive}
Ali Payani and Faramarz Fekri.
\newblock Inductive logic programming via differentiable deep neural logic
  networks.
\newblock {\em CoRR}, abs/1906.03523, 2019.

\bibitem[\protect\citeauthoryear{Riegel \bgroup \em et al.\egroup
  }{2020}]{riegel2020logical}
Ryan Riegel, Alexander~G. Gray, Francois P.~S. Luus, Naweed Khan, Ndivhuwo
  Makondo, Ismail~Yunus Akhalwaya, Haifeng Qian, Ronald Fagin, Francisco
  Barahona, Udit Sharma, Shajith Ikbal, Hima Karanam, Sumit Neelam, Ankita
  Likhyani, and Santosh~K. Srivastava.
\newblock Logical neural networks.
\newblock {\em CoRR}, abs/2006.13155, 2020.

\bibitem[\protect\citeauthoryear{Ruder}{2016}]{ruder2016overview}
Sebastian Ruder.
\newblock An overview of gradient descent optimization algorithms.
\newblock {\em arXiv preprint arXiv:1609.04747}, 2016.

\bibitem[\protect\citeauthoryear{Sen \bgroup \em et al.\egroup
  }{2022}]{sen2021neurosymbolic}
Prithviraj Sen, Breno~WSR de~Carvalho, Ryan Riegel, and Alexander Gray.
\newblock Neuro-symbolic inductive logic programming with logical neural
  networks.
\newblock In {\em Proceedings of the AAAI Conference on Artificial
  Intelligence}, volume~36, pages 8212--8219, 2022.

\bibitem[\protect\citeauthoryear{Shindo \bgroup \em et al.\egroup
  }{2021}]{Shindo_Nishino_Yamamoto_2021}
Hikaru Shindo, Masaaki Nishino, and Akihiro Yamamoto.
\newblock Differentiable inductive logic programming for structured examples.
\newblock {\em Proceedings of the AAAI Conference on Artificial Intelligence},
  35(6):5034--5041, May 2021.

\bibitem[\protect\citeauthoryear{Sitt{\'o}n \bgroup \em et al.\egroup
  }{2019}]{sitton2019neuro}
In{\'e}s Sitt{\'o}n, Ricardo~S Alonso, Elena Hern{\'a}ndez-Nieves, Sara
  Rodr{\'\i}guez-Gonzalez, and Alberto Rivas.
\newblock Neuro-symbolic hybrid systems for industry 4.0: A systematic mapping
  study.
\newblock In {\em International Conference on Knowledge Management in
  Organizations}, pages 455--465. Springer, 2019.

\bibitem[\protect\citeauthoryear{Xu \bgroup \em et al.\egroup }{2018}]{Xu18}
Jingyi Xu, Zilu Zhang, Tal Friedman, Yitao Liang, and Guy Broeck.
\newblock A semantic loss function for deep learning with symbolic knowledge.
\newblock In {\em International conference on machine learning}, pages
  5502--5511. PMLR, 2018.

\bibitem[\protect\citeauthoryear{Yang \bgroup \em et al.\egroup
  }{2020}]{neurasp}
Zhun Yang, Adam Ishay, and Joohyung Lee.
\newblock Neurasp: Embracing neural networks into answer set programming.
\newblock In Christian Bessiere, editor, {\em Proceedings of the Twenty-Ninth
  International Joint Conference on Artificial Intelligence, {IJCAI-20}}, pages
  1755--1762. International Joint Conferences on Artificial Intelligence
  Organization, 7 2020.

\end{thebibliography}

\clearpage
\appendix
\section{Appendix}

\subsection{Experiment Setup}

\subsubsection{Relying on neural network confidence for corrective example weights}\label{sec:reducing_lambda}

\begin{figure}[t]
\centering
\begin{subfigure}[t]{\linewidth}
    \centering
    \includegraphics[width=1\linewidth]{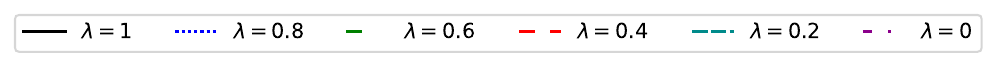}
\end{subfigure}
\begin{subfigure}[b]{0.48\linewidth}
    \centering
    \includegraphics[width=\linewidth]{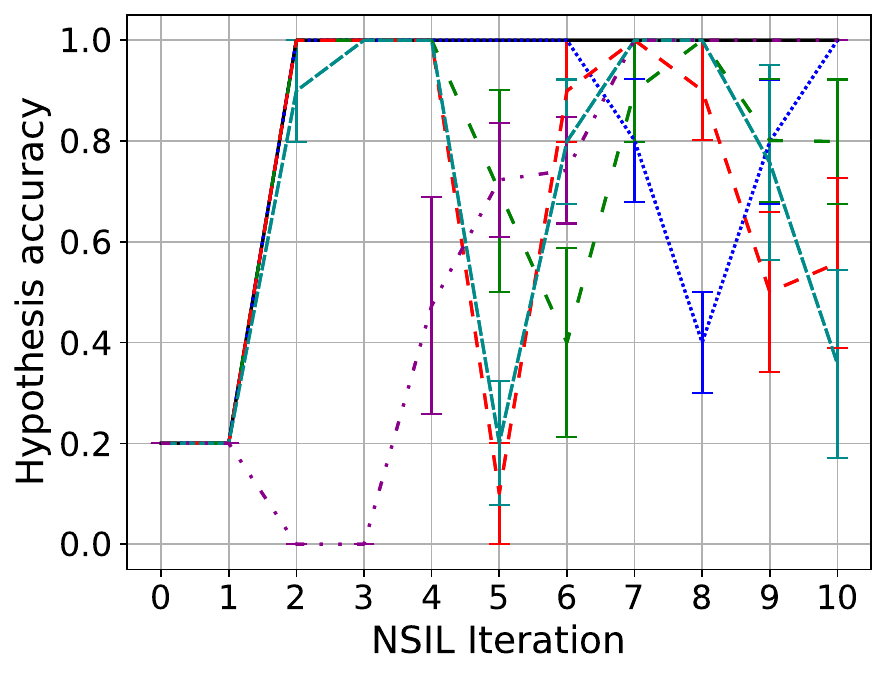}
    \caption{E9P}
\end{subfigure}
\begin{subfigure}[b]{0.48\linewidth}
    \centering
    \includegraphics[width=\linewidth]{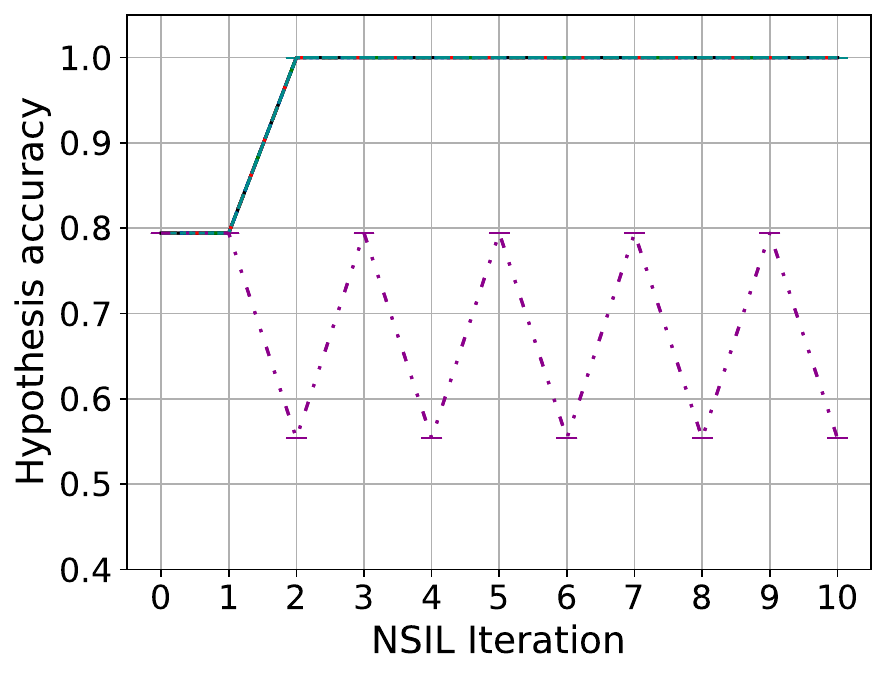}
    \caption{HS MNIST}
\end{subfigure}

\begin{subfigure}[b]{\linewidth}
    \centering
    \includegraphics[width=0.48\linewidth]{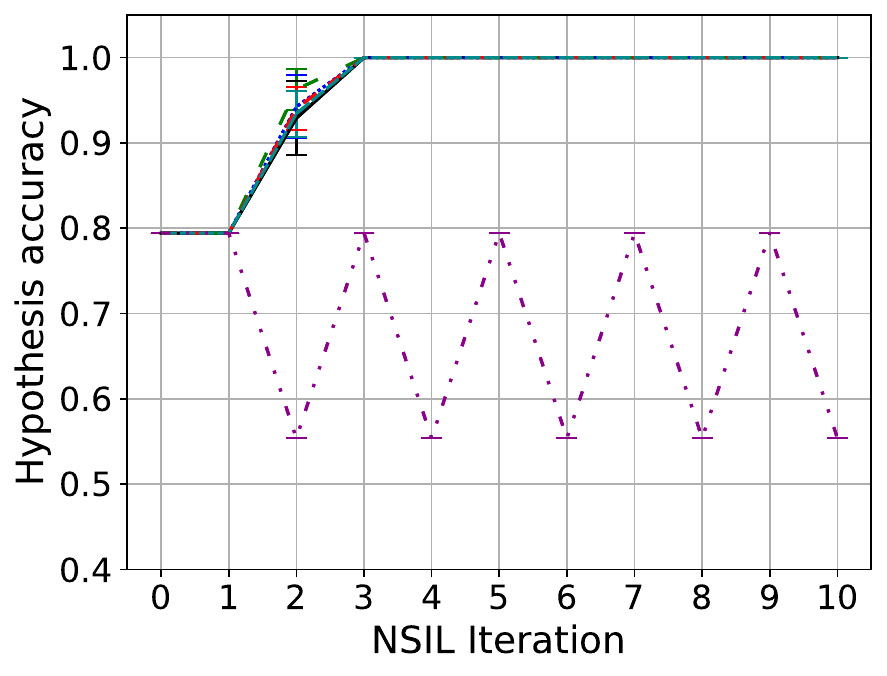}
    \caption{HS FashionMNIST}
\end{subfigure}

\caption{Knowledge accuracy with varying $\lambda$, E9P and HS tasks. Error bars indicate standard error.}
\label{fig:varying_lambda}
\end{figure}

We investigate the $\lambda$ hyper-parameter for \ac{nsl} introduced in Section \ref{sec:method:corrective_ex} and its effect on the optimisation of the symbolic learner. When \ac{nsl} calculates the weights of the corrective examples, two updates occur; one based on $FNR_{H^{\prime},\theta^{\prime}}$, and the other based on $CONF_{\theta^{\prime}}$. $\lambda$ can be adjusted to vary the effect of the $CONF_{\theta^{\prime}}$ update in Equations \ref{eq:nn_pos_weights} and \ref{eq:nn_neg_weights}, as we may not want to fully rely on the neural network confidence when the neural network is under-trained. Therefore, we use $\lambda \in \{1, 0.8, 0.6, 0.4, 0.2, 0\}$ and record the accuracy of the learned knowledge at each iteration. We again select the more challenging E9P and HS tasks, and use 40\% of the training data in the E9P task. We run \ac{nsl} for 10 iterations, with 5 repeats, using different randomly generated seeds. Figure~\ref{fig:varying_lambda} presents the results. 

In all tasks, setting $\lambda=1$ results in the best performance, as \ac{nsl} converges to the correct knowledge. This is our justification for using $\lambda=1$ in Section \ref{sec:method:corrective_ex}. In the E9P task, when $\lambda=0$ (purple dash-dot line), the accuracy improves throughout training and the symbolic learner converges to the correct knowledge at iteration 7, although has to explore more of the search space to do so. When $0 < \lambda < 1$, accurate knowledge is learned until iteration 4, at which point the accuracy begins to fluctuate. In these cases, the corrective examples with accurate neural network predictions have lower weights, which encourages the symbolic learner to choose a different label than the label given in the training set. When $\lambda=1$, \ac{nsl} maintains the correct knowledge in later stages of training. In both HS tasks, $0 < \lambda \leq 1$ results in similar performance, whereas when $\lambda=0$, the learned knowledge alternates at each iteration. This is due to the ``broad-brush" nature of the \ac{fnr} in these tasks, as many examples have the same weight and are updated simultaneously. There are only two possible labels for each set configuration, and there are many choices of digits, so the \ac{fnr} for each label and set configuration applies to many corrective examples. Note this is not the case in the E9P task, as there are 19 labels and fewer choices of digits for each label. In the HS tasks, when $\lambda>0$, the neural network confidence scores help the symbolic learner to converge to the correct knowledge by providing a positive weight update for specific examples. This breaks down the broad weights set by the \ac{fnr}.

\subsubsection{Domain knowledge and corrective examples}\label{sec:domain_knowledge}
\ac{asp} encodings for the domain knowledge are presented in Figure \ref{fig:bk}. Let us now present sample encodings for the corrective examples. In each case, $\asp{ID}$ refers to a unique identifier, and $\asp{W}$ is the example weight calculated by Equations \ref{eq:pi_y_pos_weight} - \ref{eq:nn_neg_weights}.

\noindent \textbf{Cumulative Arithmetic} Let us assume the Addition sub-task, $X=[\inlinegraphics{figures/digits/1.png}, \inlinegraphics{figures/digits/3.png}, \inlinegraphics{figures/digits/4.png}]$, $y=8$, and $Z=[1,3,4]$. A corrective example pair for this data point is shown in Figures \ref{fig:recursive_example_pos} and \ref{fig:recursive_example_neg}.

\noindent \textbf{Two-Digit Arithmetic} Let us assume the E9P sub-task, $X=[\inlinegraphics{figures/digits/2.png}, \inlinegraphics{figures/digits/1.png}]$, $y=1$, and $Z=[2,1]$. A corrective example pair for this data point is shown in Figures \ref{fig:e9p_example_pos} and \ref{fig:e9p_example_neg}.

\noindent \textbf{Hitting Set} Let us assume the HS sub-task, $X= \left \{ \{ \inlinegraphics{figures/digits/1.png} \}, \{\inlinegraphics{figures/digits/1_1.png}, \inlinegraphics{figures/digits/3.png} \}, \{ \inlinegraphics{figures/digits/4.png}\} \right \}$, $y=1$, and $Z= \left \{ \{ 1 \}, \{1, 3 \}, \{ 4\} \right \}$. A corrective example pair for this data point is shown in Figures \ref{fig:hs_example_pos} and \ref{fig:hs_example_neg}. Note that $e^{\text{pos}}_{Z,y}$ and $e^{\text{neg}}_{Z,y}$ are implemented with positive and negative \ac{las} examples respectively.

\begin{figure}
    \centering
    \begin{subfigure}[m]{0.30\textwidth}
        \begin{lstlisting}
        #pos(ID@W, { result(8) }, {  }, {
            :- result(X), X != 8.
            start_list((4, (3, (1, empty)))).
        }).
        \end{lstlisting}
        \caption{Cumulative Addition $e^{\text{pos}}_{Z,y}$}
        \label{fig:recursive_example_pos}
    \end{subfigure}
    \begin{subfigure}[m]{0.30\textwidth}
        \begin{lstlisting}
        #pos(ID@W, { result }, {  }, {
            result :- result(X), X != 8.
            :- result(X), result(Y), Y < X.
            start_list((4, (3, (1, empty)))).
        }).
        \end{lstlisting}
        \caption{Cumulative Addition $e^{\text{neg}}_{Z,y}$}
        \label{fig:recursive_example_neg}
    \end{subfigure}
    
    \vspace{2em}
    \begin{subfigure}[m]{0.30\textwidth}
        \begin{lstlisting}
        #pos(ID@W, { result(1) }, {  }, {
            :- result(X), X != 1.
            digit(1, 2). digit(2, 1).
        }).
        \end{lstlisting}
        \caption{E9P $e^{\text{pos}}_{Z,y}$}
        \label{fig:e9p_example_pos}
    \end{subfigure}
    \begin{subfigure}[m]{0.30\textwidth}
        \begin{lstlisting}
        #pos(ID@W, { result }, {  }, {
            result :- result(X), X != 1.
            :- result(X), result(Y), Y < X.
            digit(1, 2). digit(2, 1).
        }).
        \end{lstlisting}
        \caption{E9P $e^{\text{neg}}_{Z,y}$}
        \label{fig:e9p_example_neg}
    \end{subfigure}
    
    \vspace{2em}
    \begin{subfigure}{0.30\textwidth}
        \begin{lstlisting}
        #pos(ID@W, { }, {  }, {
            ss_element(1,1).
            ss_element(2,1).
            ss_element(2,3).
            ss_element(3,4).
        }).
        \end{lstlisting}
        \caption{HS $e^{\text{pos}}_{Z,y}$}
        \label{fig:hs_example_pos}
    \end{subfigure}
    \begin{subfigure}{0.30\textwidth}
        \begin{lstlisting}
        #neg(ID@W, { }, {  }, {
            ss_element(1,1).
            ss_element(2,1).
            ss_element(2,3).
            ss_element(3,4).
        }).
        \end{lstlisting}
        \caption{HS $e^{\text{neg}}_{Z,y}$}
        \label{fig:hs_example_neg}
    \end{subfigure}
\end{figure}

\begin{figure}
    \centering
    \begin{subfigure}{0.90\linewidth}
        \begin{lstlisting}
empty(empty). 
list(L) :- start_list(L).
list(T) :- list((_, T)).
head(L, H) :- list(L), L = (H, _).
tail(L, T) :- list(L), L = (_, T).
add(L, (X+Y, T)) :- list(L), L = (X, (Y, T)).
list(L) :- add(_, L).
mult(L, (X*Y, T)) :- list(L), L = (X, (Y, T)).
list(L) :- mult(_, L).
eq(L, ELT) :- list(L), L = (ELT, empty).
result(R) :- start_list(L), f(L, R).
:- result(X), result(Y), X < Y.
#predicate(base, head/2). #predicate(base, tail/2).
#predicate(base, add/2). #predicate(base, mult/2).
#predicate(base, eq/2). #predicate(base, empty/1).
#predicate(target, f/2).
P(A, B) :- Q(A, B), m1(P, Q).
P(A, B) :- Q(A, C), P(C, B), m2(P, Q).
P(A, B) :- Q(A, C), R(C, B), m3(P, Q, R), Q != R.
P(A, B) :- Q(A, B), R(A, B), m4(P, Q, R), Q != R.
P(A) :- Q(A, B), m5(P, Q, B).
P(A) :- Q(A), m6(P, Q).
P(A, B) :- Q(A), R(A, B), m7(P, Q, R).
P(A, B) :- Q(A, B), R(B), m8(P, Q, R).
#modem(2, m1(target/2, any/2)).
#modem(2, m2(target/2, any/2)).
#modem(3, m3(target/2, any/2, any/2)).
#modem(3, m4(target/2, any/2, any/2)).
#modem(2, m5(target/1, any/2)).
#modem(2, m6(target/1, any/1)).
#modem(3, m7(target/2, any/1, any/2)).
#modem(3, m8(target/2, any/2, any/1)).
    \end{lstlisting}
        \caption{Cumulative Arithmetic tasks}
        \label{fig:mnist_recursive_arithmetic_bk}
    \end{subfigure}
    \begin{subfigure}{0.90\linewidth}
        \begin{lstlisting}
    r(0..18). d(0..9).
    even(X) :- d(X), X \\ 2 = 0.
    plus_nine(X1,X2) :- d(X1), X2=9+X1.
    res(X1,X2,Y) :- dig(1,X1), dig(2,X2).
    :- dig(1,X1),dig(2,X2),res(X1,X2,Y1),res(X1,X2,Y2), Y1 != Y2.
    #modeh(res(var(d),var(d),var(r))).
    #modeb(var(n) = var(d)).
    #modeb(var(n) = var(d) + var(d)).
    #modeb(plus_nine(var(d),var(r))).
    #modeb(even(var(d))).
    #modeb(not even(var(d))).
    \end{lstlisting}
        \caption{Two-Digit Arithmetic tasks}
        \label{fig:mnist_arithmetic_bk}
    \end{subfigure}
    \hspace{1em}
    \begin{subfigure}{0.9\linewidth}
        \begin{lstlisting}
        s(1..4). h(1..2). e(1..5).
        #modeha(hs(var(h), var(e))).
        #modeh(hit(var(s))).
        #modeb(hs(var(h), var(e)),(positive)).
        #modeb(var(e) != var(e)).
        #modeb(ss_element(var(s),var(e)),(positive)).
        #modeb(ss_element(3,var(e)),(positive)).
        #modeb(ss_element(var(s),1),(positive)).
        #modeb(hit(var(s))).
    \end{lstlisting}
        \caption{MNIST Hitting Set tasks}
        \label{fig:hitting_sets_bk}
        \end{subfigure}
        \caption{ASP encodings of \ac{nsl} domain knowledge.}
    \label{fig:bk}
\end{figure}

\subsubsection{Model details}\label{sec:model_details}
\textbf{\ac{nsl} CNN} The neural component in \ac{nsl} is the CNN architecture from \cite{deepproblog,neurasp}. It consists of an encoder with 2 convolutional layers with kernel size 5, and output sizes 6 and 16 respectively. Each convolutional layer is followed by a max pooling layer of size 2, stride 2. The encoder is followed by 3 linear layers of size 120, 84, and 10 (in the hitting set tasks, the last layer is of size 5), and all layers are followed by a ReLU activation function. Finally, a softmax layer returns the output probability distribution. 

\textbf{Baselines} The baseline CNN in the Arithmetic tasks follows the baseline CNN architecture from \cite{deepproblog,neurasp}, which is largely the same as the \ac{nsl} CNN, except the size of the last linear layer is 19, the network accepts a pair of concatenated images as input, and a log softmax layer is used to provide a classification. The CNN-LSTM in the Hitting Set tasks uses the same CNN encoder as \ac{nsl}, applied to each image in the input sequence, followed by an LSTM layer of tunable size, a linear layer of size 1, and a sigmoid layer. The CBM and CBM-S baselines have similar architectures, consisting of the same CNN as \ac{nsl}, applied to each image separately. The only difference is that the CBM variant doesn't contain a softmax layer for the CNN. In both variants, the CNN is followed by 3 linear layers where the size of the first two layers are tunable and the last layer is of size 19 in the Arithmetic tasks, and 1 in the Hitting Set tasks. In the Addition task, ReLU activation is used after the first two linear layers for the CBM and CBM-S baselines, as this resulted in improved performance. In the other tasks, no non-linearity was used. The CNN-LSTM-NALU and CNN-LSTM-NAC baselines contain the same CNN as \ac{nsl} and the other baselines, followed by an LSTM layer of tunable size, and then either an NALU or NAC layer. Finally, we use the stochastic gradient descent optimiser for all neural networks \cite{ruder2016overview}, and tune the learning rate and momentum.\footnote{Please refer to \url{https://github.com/DanCunnington/NSIL} for details of all the hyper-parameters used in our experiments.}

\subsubsection{Machine details}
All experiments are performed on the following infrastructure: RHEL 8.5 x86\_64 with Intel Xeon E5-2667 CPUs (20 cores total), and an NVIDIA A100 GPU, 150GB RAM.

\begin{figure}[t]
\centering
\begin{subfigure}[t]{0.85\linewidth}
    \centering
    \includegraphics[width=\textwidth]{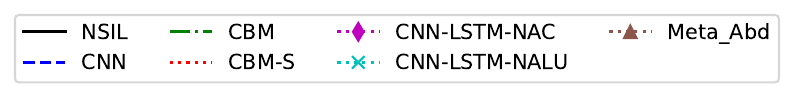}
\end{subfigure}
\begin{subfigure}[t]{0.49\linewidth}
    \includegraphics[width=\linewidth]{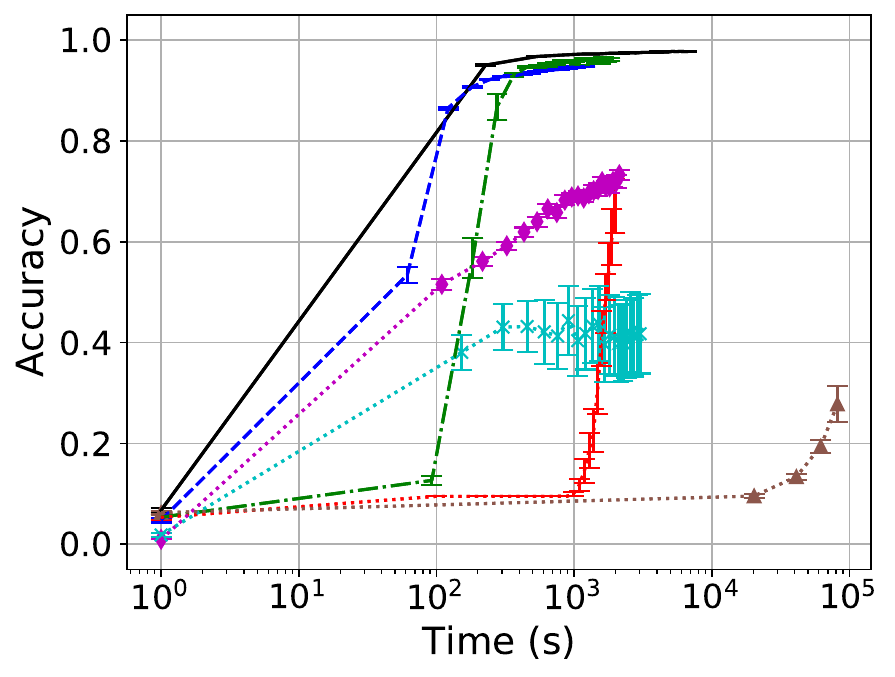}
    \caption{Addition 100\%}
\end{subfigure}
\begin{subfigure}[t]{0.49\linewidth}
    \includegraphics[width=\linewidth]{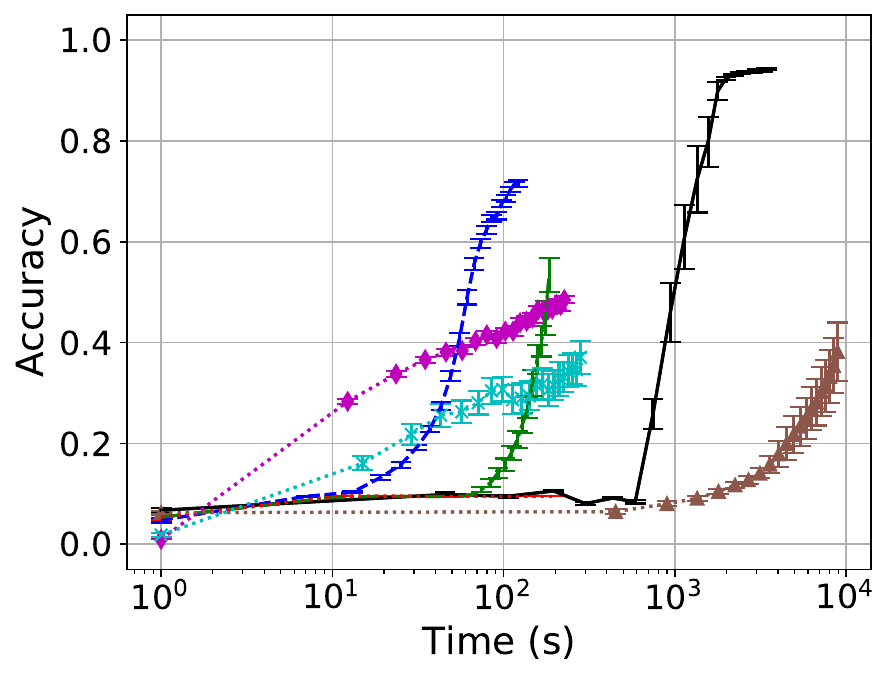}
    \caption{Addition 10\%}
\end{subfigure}
\begin{subfigure}[t]{0.49\linewidth}
    \includegraphics[width=\linewidth]{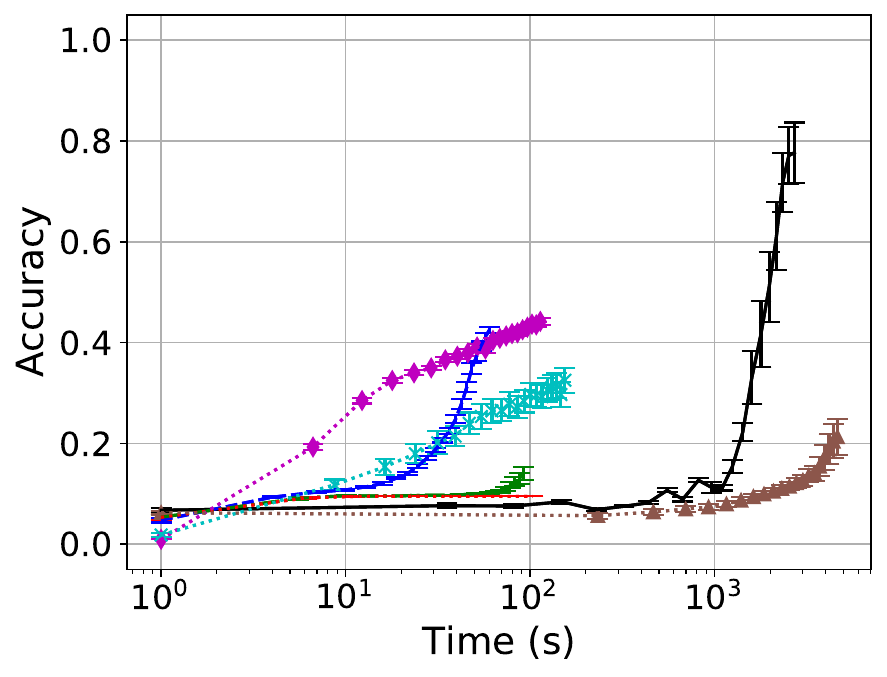}
    \caption{Addition 5\%}
\end{subfigure}
\begin{subfigure}[t]{0.49\linewidth}
    \includegraphics[width=\linewidth]{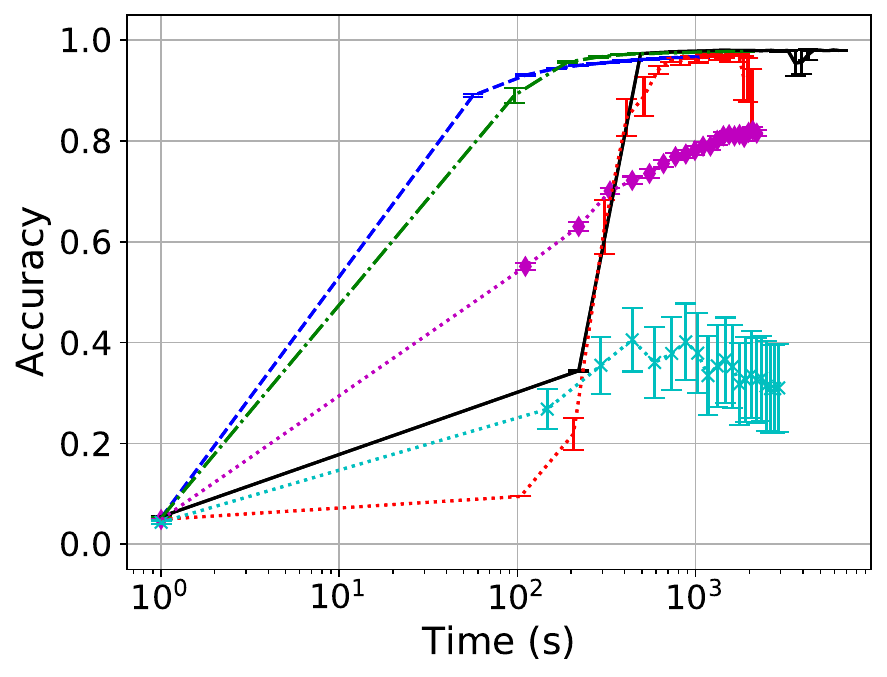}
    \caption{E9P 100\%}
\end{subfigure}
\begin{subfigure}[t]{0.49\linewidth}
    \includegraphics[width=\linewidth]{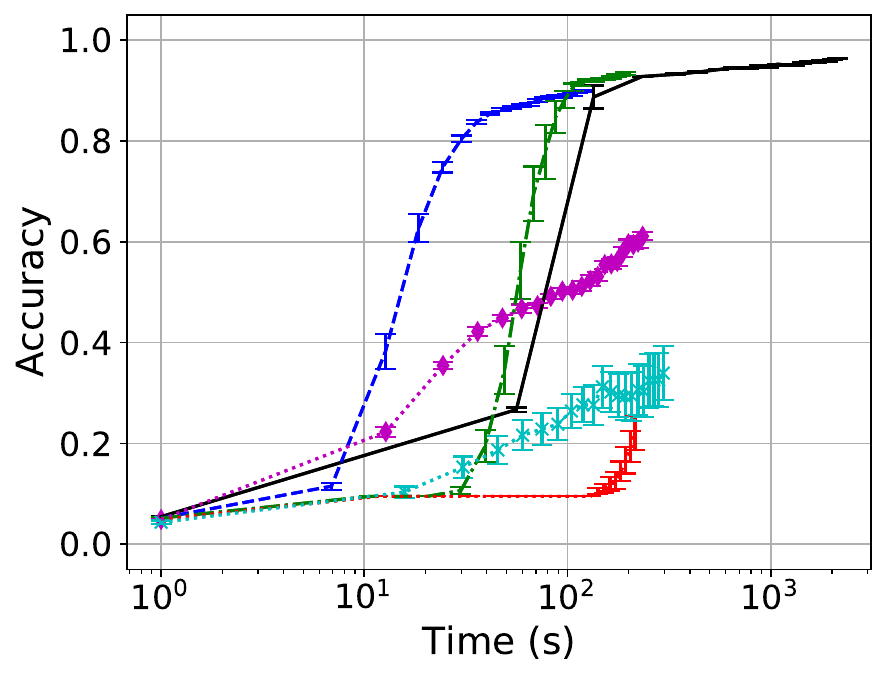}
    \caption{E9P 10\%}
\end{subfigure}
\begin{subfigure}[t]{0.49\linewidth}
    \includegraphics[width=\linewidth]{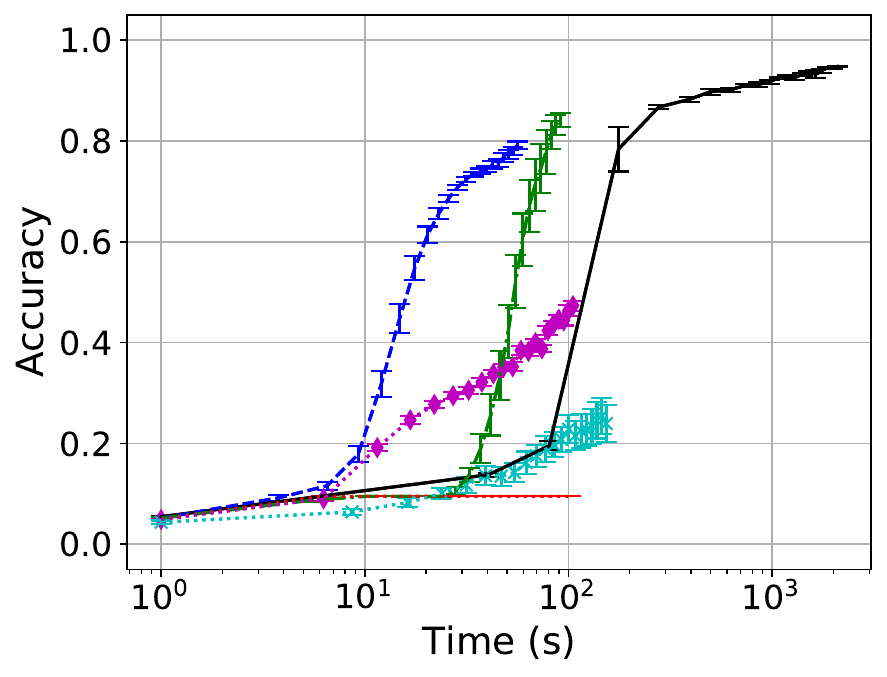}
    \caption{E9P 5\%}
\end{subfigure}
\caption{Two-Digit Arithmetic learning time vs. accuracy with reducing training set sizes.}
\label{fig:mnist_arithmetic_time}
\end{figure}

\begin{figure}[t]
\centering
\begin{subfigure}[t]{1\linewidth}
    \centering
    \includegraphics[width=0.75\linewidth]{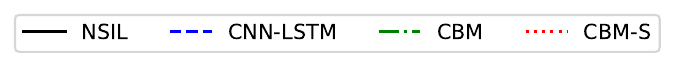}
\end{subfigure}
\begin{subfigure}[t]{0.49\linewidth}
    \includegraphics[width=\linewidth]{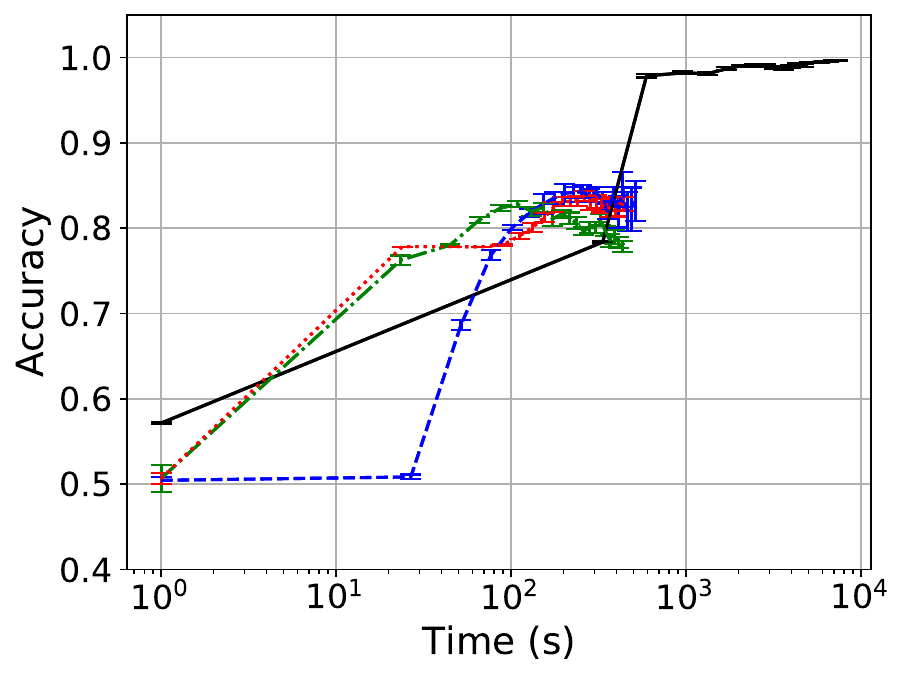}
    \caption{HS MNIST}
\end{subfigure}
\begin{subfigure}[t]{0.49\linewidth}
    \includegraphics[width=\linewidth]{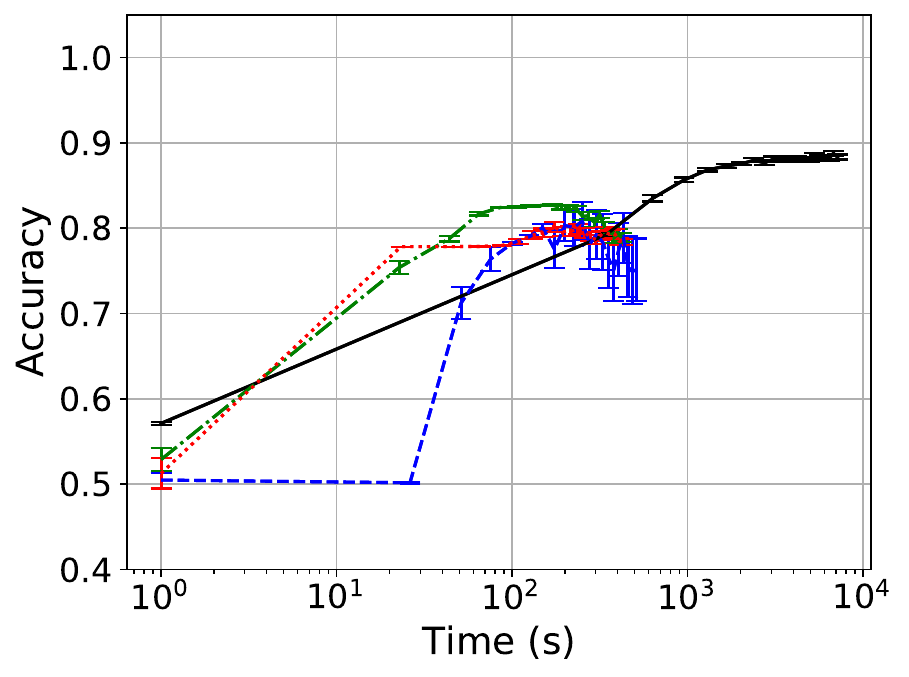}
    \caption{HS FashionMNIST}
\end{subfigure}
\begin{subfigure}[t]{0.49\linewidth}
    \includegraphics[width=\linewidth]{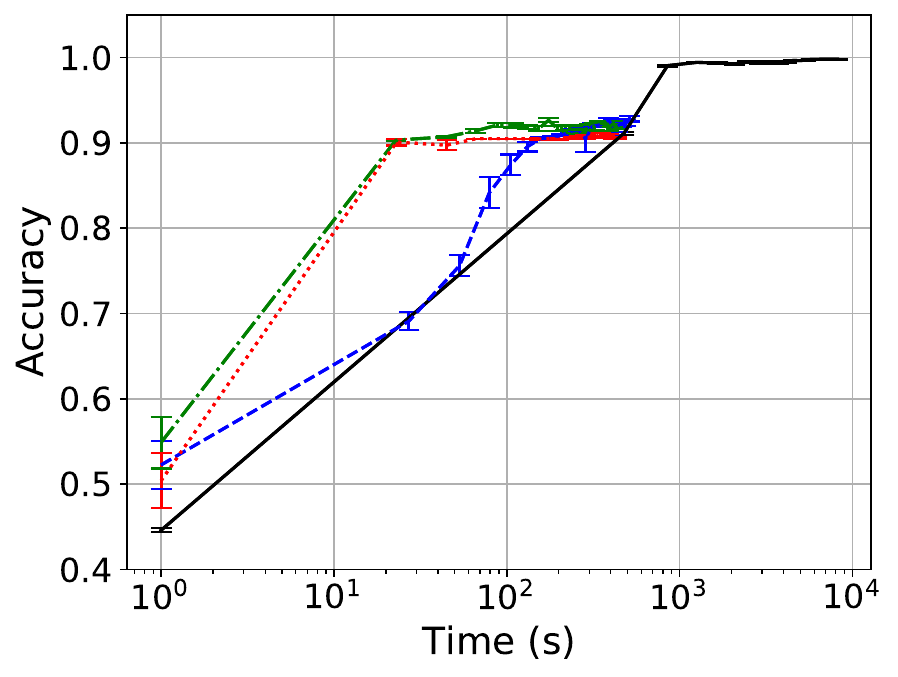}
    \caption{CHS MNIST}
\end{subfigure}
\begin{subfigure}[t]{0.49\linewidth}
    \includegraphics[width=\linewidth]{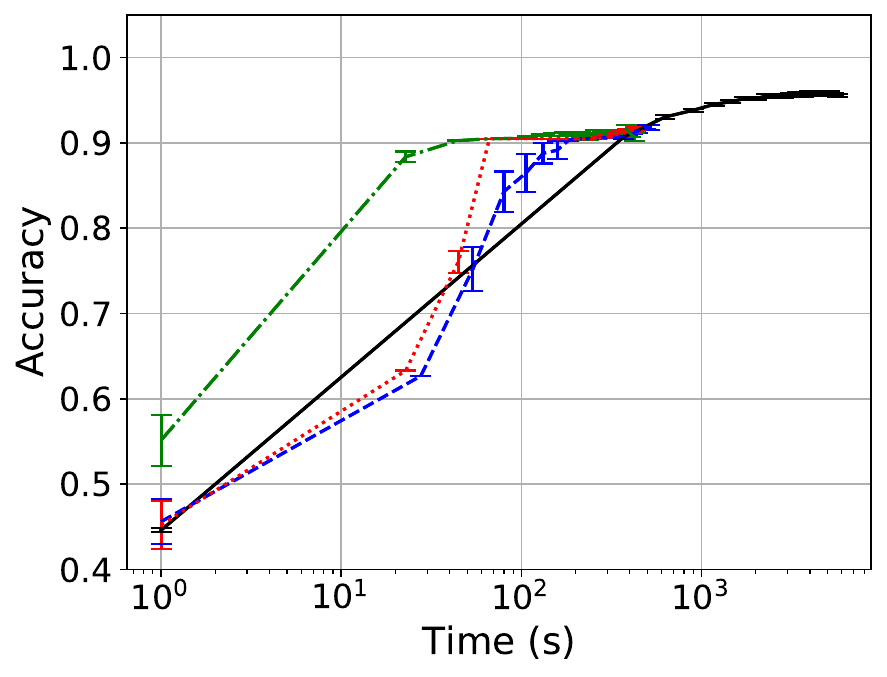}
    \caption{CHS FashionMNIST}
\end{subfigure}

\caption{Hitting Set learning time vs. accuracy.}
\label{fig:hitting_sets_time}
\end{figure}

\subsection{Learning Time Comparison}\label{sec:run_time_comparison}
Figures~\ref{fig:mnist_arithmetic_time} and~\ref{fig:hitting_sets_time} show the learning time vs. accuracy comparison for the Arithmetic and Hitting Set tasks respectively. For each method we plot the time taken to complete 20 epochs, and each point shows the accuracy after an epoch of training (1 epoch = 1 \ac{nsl} iteration). Error bars indicate standard error. As observed in both task domains, \ac{nsl} requires more time to complete 20 epochs, but achieves a greater accuracy than the neural methods. Finally, \ac{nsl} has comparable learning time to $Meta_{Abd}$ on the addition tasks.

\subsection{Asset Licenses}
The ILASP system is free to use for research,\footnote{\url{https://ilasp.com/terms}} 
 FastLAS\footnote{\url{https://github.com/spike-imperial/FastLAS/blob/master/LICENSE}} and the FashionMNIST dataset\footnote{\url{https://github.com/zalandoresearch/fashion-mnist/blob/master/LICENSE}} are both open-source with an MIT license, the MNIST dataset is licensed with Creative Commons Attribution-Share Alike 3.0,\footnote{See bottom paragraph: \url{https://keras.io/api/datasets/mnist/}} and the CNN models used from DeepProbLog are open-source and licensed with Apache 2.0.\footnote{\url{https://github.com/ML-KULeuven/deepproblog/blob/master/LICENSE}}

\end{document}